\documentclass{article} % For LaTeX2e

\usepackage[utf8]{inputenc}
\usepackage{algorithm}
\usepackage{algpseudocode}
\usepackage{amsmath}
\usepackage{amssymb} 
\usepackage{graphicx}

\usepackage[T1]{fontenc}    % use 8-bit T1 fonts
\usepackage{hyperref}       % hyperlinks
\usepackage{url}            % simple URL typesetting
\usepackage{booktabs}       % professional-quality tables
\usepackage{amsfonts}       % blackboard math symbols
\usepackage{nicefrac}       % compact symbols for 1/2, etc.
\usepackage{microtype}      % microtypography
\usepackage{xcolor}         % colors
\usepackage{amsmath,amsfonts,bm}

\def\eqref#1{equation~\ref{#1}}
\def\1{\bm{1}}

\DeclareMathAlphabet{\mathsfit}{\encodingdefault}{\sfdefault}{m}{sl}
\SetMathAlphabet{\mathsfit}{bold}{\encodingdefault}{\sfdefault}{bx}{n}

\usepackage{enumitem}
\usepackage{hyperref}
\usepackage{url}
\usepackage[preprint]{neurips_2024}
\usepackage{colortbl}
\usepackage{multirow}
\usepackage{array}
\usepackage{xspace}
\usepackage{subcaption}
\usepackage{makecell}
\usepackage{pifont}

\title{Knocking-Heads Attention}
\author{
  Zhanchao Zhou$^{1,2,3}$, 
  Xiaodong Chen$^{1,4}$, 
  Haoxing Chen$^{1}$, 
  Zhenzhong Lan$^{1,3\dagger}$, 
  Jianguo Li$^{1\dagger}$
}
\affiliation{
  $^1$ Ant Group \quad
  $^2$ Zhejiang University \quad
  $^3$ Westlake University \quad
  $^4$ Renmin University of China \quad
}

\newcommand{\eg}{\emph{e.g.,}\xspace}
\newcommand{\vsc}{\emph{vs.}\xspace}
\newcommand{\ie}{\emph{i.e.}\xspace}
\definecolor{darkgreen}{RGB}{0,100,0}
\definecolor{darkred}{RGB}{139,0,0}
\begin{document}
\maketitle

\begin{abstract}
Multi-head attention (MHA) has become the cornerstone of modern large language models, enhancing representational capacity through parallel attention heads. However, increasing the number of heads inherently weakens individual head capacity, and existing attention mechanisms — whether standard MHA or its variants like grouped-query attention (GQA) and grouped-tied attention (GTA) — simply concatenate outputs from isolated heads without strong interaction. To address this limitation, we propose knocking-heads attention (KHA), which enables attention heads to ``knock" on each other — facilitating cross-head feature-level interactions before the scaled dot-product attention. This is achieved by applying a shared, diagonally-initialized projection matrix across all heads. The diagonal initialization preserves head-specific specialization at the start of training while allowing the model to progressively learn integrated cross-head representations. KHA adds only minimal parameters and FLOPs and can be seamlessly integrated into MHA, GQA, GTA, and other attention variants. We validate KHA by training a 6.1B parameter MoE model (1.01B activated) on 1T high-quality tokens. Compared to baseline attention mechanisms, KHA brings superior and more stable training dynamics, achieving better performance across downstream tasks.
\end{abstract}
\begin{figure}[!b]
\vspace{-7mm}
\centering
    \begin{minipage}{0.47\textwidth}
        \centering
        \includegraphics[width=\textwidth, trim={5cm 5cm 7.1cm 4cm}, clip]{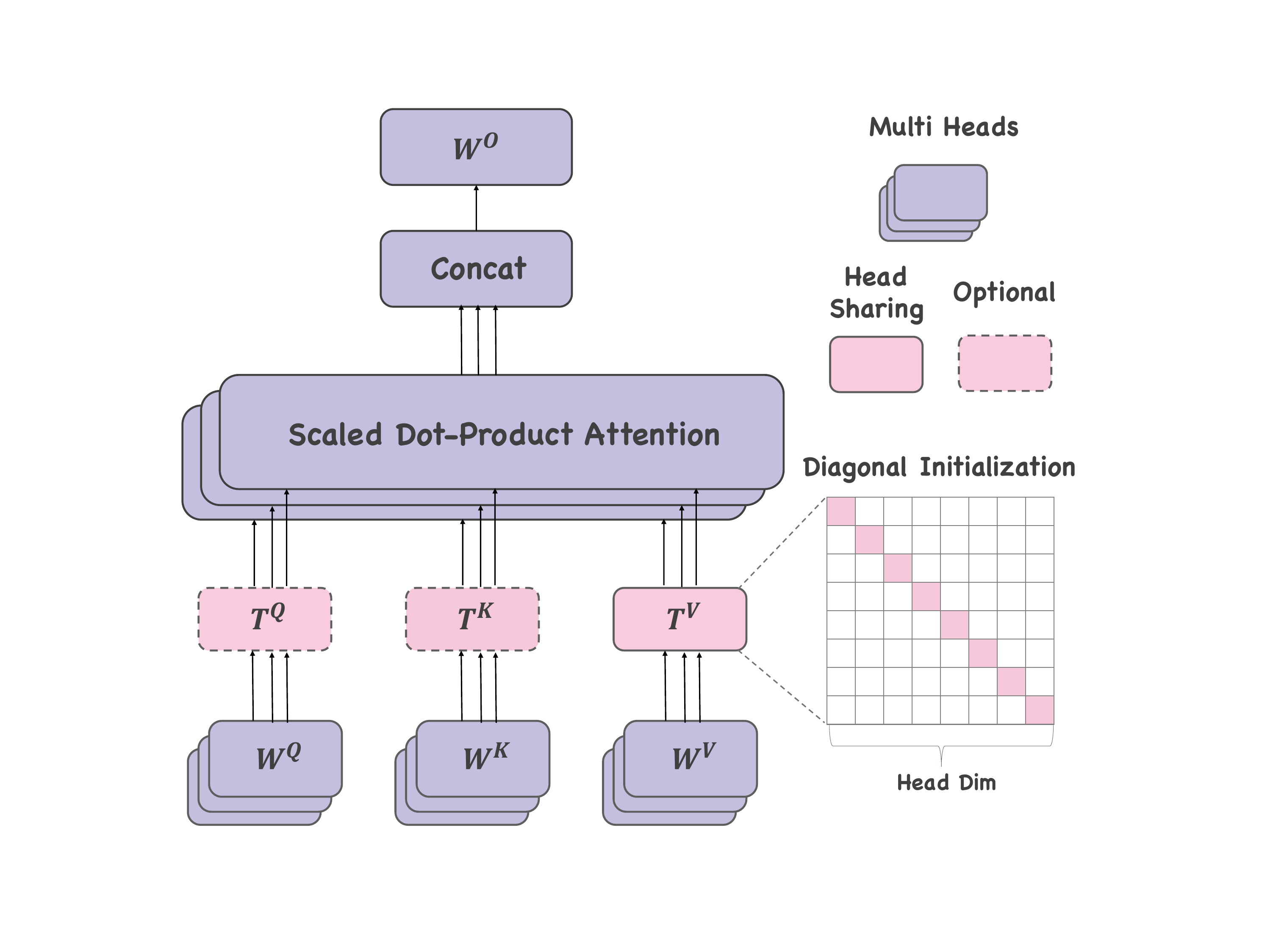}
    \end{minipage}
    \hspace{-3mm}
    \begin{minipage}{0.51\textwidth}
        \centering
        \includegraphics[width=0.85\textwidth]{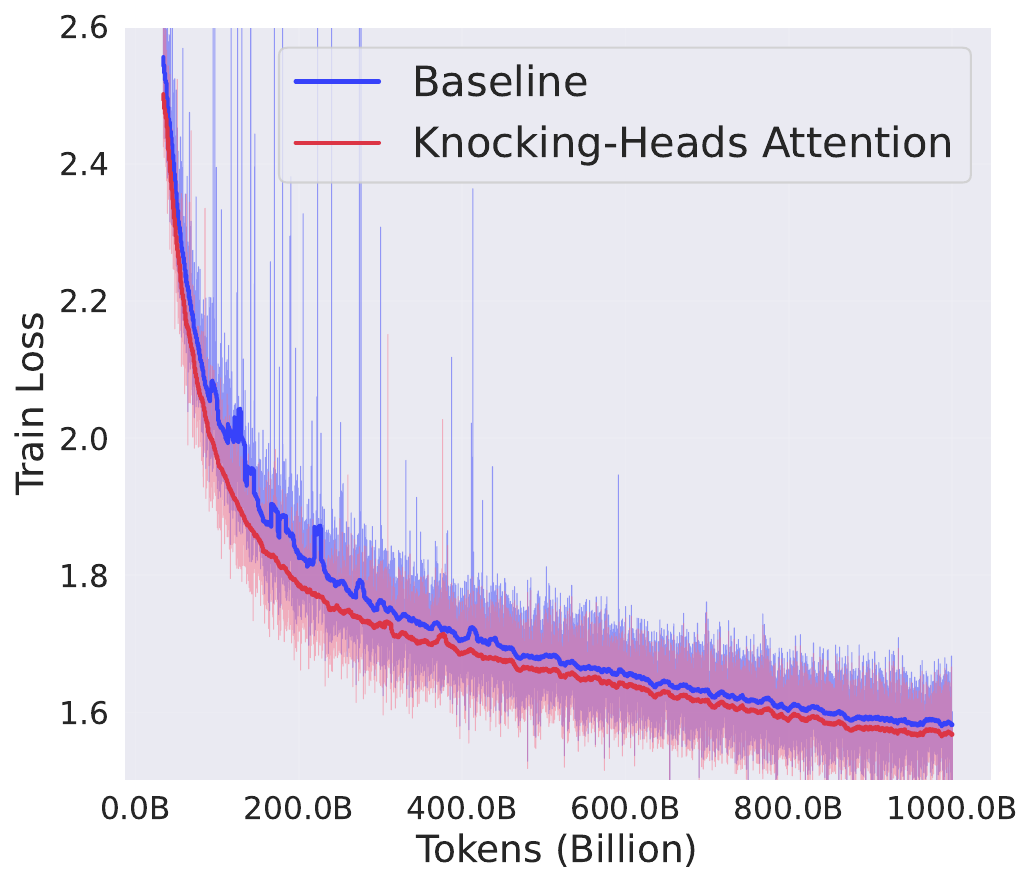}
    \end{minipage}
    \caption{(Left) The knocking-heads attention architecture. Purple represents the original multi-head attention, while pink represents the added knocking-heads projections. $T^Q$ and $T^K$ within the dashed box are optional projections due to their lower importance compared to $T^V$. (Right) Training loss over 1T tokens for 6.1B MoE models (1.01B activated parameters): baseline \vsc knocking-heads attention. KHA reduces loss spikes and maintains consistently lower training loss.}
    \label{fig:main}
\end{figure}
\section{Introduction}
One of the key factors behind the success of large language models is the design of multi-head attention \citep{vaswani2017attention}, which has been preserved across various subsequent attention variants \citep{shazeer2020talking,ainslie2023gqa,liu2024deepseek,zadouri2025hardware}. Multi-head attention (MHA) enables models to simultaneously process information from multiple representation subspaces across different sequence positions, with each attention head serving distinct functions in sequence modeling \citep{xiao2023efficient,xiao2024duoattention,qin2025elucidating}.

However, different heads operate independently when modeling attention, with each head computing its output separately before concatenation and forward propagation, lacking mutual interaction. Talking-heads attention \citep{shazeer2020talking} addresses this limitation by introducing linear projections on either logits before the attention softmax operation or attention scores after the softmax operation to enable inter-head communication. Nevertheless, talking-heads projections on quadratically-scaling attention matrices lead to dramatically increased computational complexity, especially when the number of heads is large.

In this paper, we propose knocking-heads attention (KHA) (Fig.~\ref{fig:main} (Left)), a variant of multi-head attention that achieves inter-head interaction at the feature level through unified transformations applied to all heads. Compared to standard multi-head attention, KHA introduces additional shared projections for query, key, and value representations, supplementing the original head-specific projections. Among these, the knocking-heads projections for queries and keys are optional since they provide smaller improvements compared to the value projections. We initialize these knocking-heads projections with diagonal matrices, allowing the model to start from isolated heads and gradually develop inter-head communication during training, eventually reaching an optimal balance between head specialization and cross-head collaboration. This additional head-sharing mechanism not only facilitates information sharing and coordination between different heads while preserving their individual specializations, but also serves as an implicit regularization constraint applied to all heads. Knocking-heads projections introduce minimal overhead ($<1$\% additional FLOPs and parameters) during training and, when implemented as linear transformations rather than MLPs, can be fully integrated into the original projections at inference time.

We conduct extensive experiments on language modeling. We first perform architectural exploration to identify the optimal configuration for knocking-heads projections, examining different projection types (linear, MLP-based, gated), target components (queries, keys, values), attention variants (MHA, MQA, GQA, GTA), and head configurations. Our analysis reveals that applying MLP-based knocking-heads projections to values yields the most significant improvements, with the benefits becoming evident at 4 value/key heads. Based on these findings, we adopt GQA with 32 query heads and 4 KV groups, enabling knocking-heads projections to achieve high performance while maintaining efficient KV caching. We validate our approach through comprehensive experiments on 1T tokens using 1.01B-active-parameter, 6.1B-total-parameter MoE models, with results shown in Fig.~\ref{fig:main} (Right). Our method achieves significant improvements in Language Understanding (+4.32), Code (+3.9), and Math (+1.62) tasks, with an overall average score improvement of 1.26 points across all evaluated tasks. Additional experiments across various MoE and dense model scales further demonstrate the effectiveness of our method. We summarize our main contributions as follows:

\begin{itemize}
\item We introduce knocking-heads attention, a novel head interaction mechanism that enables cross-head communication through shared transformation matrices with minimal computational overhead. To preserve head specialization, we propose diagonal initialization that allows heads to maintain their distinct roles while gradually developing beneficial cross-head interactions.
\item We conduct extensive experiments validating KHA's universality across different attention mechanisms, model types (MoE and dense), and model scales, demonstrating broad applicability and identifying optimal configurations.
\item We demonstrate scalability through large-scale experiments on 1T tokens using 1.01B-active-parameter, 6.1B-total-parameter models, achieving consistent improvements in training stability and model performance.
\end{itemize}

\section{Related Work}
\subsection{Parameter Sharing in Architecture Design}
Parameter sharing is a fundamental design principle in deep learning architectures. In CNNs \citep{lecun2002gradient}, shared convolutional kernels enable translation equivariance and improve generalization capability. ALBERT \citep{lan2019albert} achieves significant parameter reduction through cross-layer parameter sharing. DeepSeek-MoE \citep{dai2024deepseekmoe} introduces shared experts to reduce parameter redundancy across routed experts. In this work, we introduce additional shared projection matrices (\ie, knocking-heads projections) across different attention heads on top of the original head-specific projections to facilitate inter-head interaction and provide regularization effects.

\subsection{Attention Head Interaction}
Multi-head attention mechanisms allow different attention heads to learn diverse patterns~\citep{xiao2023efficient,xiao2024duoattention}, but these heads lack interconnection and often exhibit significant redundancy~\citep{cordonnier2020multi,jin2024moh}. Several approaches have explored inter-head communication to address this issue. Talking-heads attention~\citep{shazeer2020talking} applies learnable transformations to attention weight matrices, but requires materializing these matrices, making it incompatible with FlashAttention~\citep{dao2022flashattention}, not to mention its substantial computational overhead when head counts are large relative to head dimensions. Collaborated multi-head attention~\citep{cordonnier2020multi} shares projection matrices across heads with head-wise dimension-specific mixers, yet increases training FLOPs while limiting head specialization to feature reweighting. Mixture-of-head attention~\citep{jin2024moh} dynamically selects head subsets per token but provides limited inter-head communication while incurring complex router training. Our proposed knocking-heads attention adopts head-sharing projections similar to collaborated attention but preserves head specialization through keeping original head-specific projections and diagonal initialization. The plug-and-play mechanism is compatible with any attention variants and FlashAttention framework, adding minimal FLOPs and parameters. More detailed comparisons are provided in Table \ref{tab:comparison} in the appendix.

\subsection{Loss Spikes in Pre-training}
Loss spikes are a common challenge in large-scale pretraining. Recent work has identified various underlying causes and solutions beyond simply skipping problematic data batches \citep{chowdhery2023palm}. \cite{takase2023spike} found correlations between loss spikes and gradient spikes in specific parameters on 1.7B parameter models, proposing scaled initialization for embedding layers. \cite{qiu2025gated} reduced loss spikes by introducing gating mechanisms to avoid massive activations. Kimi-K2 \citep{team2025kimi} identified attention logit spikes in certain heads as the primary cause and proposed QK-clip for mitigation. Our knocking-heads structure addresses this issue differently—by sharing transformation matrices across attention heads, it introduces regularization effects that effectively mitigate loss spikes.

\section{Method}
\subsection{Classical Attention}
Multi-head attention (MHA) enables the model to jointly attend to information from different representation subspaces. Given an input sequence $\mathbf{X} \in \mathbb{R}^{L \times d}$ where $L$ is the sequence length and $d$ is the hidden dimension, MHA employs $n$ parallel attention heads to capture diverse attention patterns.

For each attention head $i \in \{1, 2, \ldots, n\}$, the queries, keys, and values are computed as:
\begin{align}
\mathbf{Q}_i &= \mathbf{X} \mathbf{W}_i^Q, \quad \mathbf{W}_i^Q \in \mathbb{R}^{d \times d_k}, \\
\mathbf{K}_i &= \mathbf{X} \mathbf{W}_i^K, \quad \mathbf{W}_i^K \in \mathbb{R}^{d \times d_k}, \\
\mathbf{V}_i &= \mathbf{X} \mathbf{W}_i^V, \quad \mathbf{W}_i^V \in \mathbb{R}^{d \times d_v},
\end{align}
where $d_k = d_v = d/n$ represents the dimension of each head. The attention output for head $i$ is computed as $\mathbf{O}_i = \text{Attention}(\mathbf{Q}_i, \mathbf{K}_i, \mathbf{V}_i) = \text{Softmax}\left(\frac{\mathbf{Q}_i \mathbf{K}_i^T}{\sqrt{d_k}}\right) \mathbf{V}_i$, following the scaled dot-product attention mechanism. The final MHA output concatenates all head outputs and applies a linear projection:
\begin{equation}
\text{MHA}(\mathbf{X}) = \text{Concat}(\mathbf{O}_1, \mathbf{O}_2, \ldots, \mathbf{O}_n) \mathbf{W}^O,
\end{equation}
where $\mathbf{W}^O \in \mathbb{R}^{d \times d}$ is the output projection matrix. Group Query Attention (GQA) extends this framework by partitioning $n$ query heads into $g$ groups, where each group shares the same key-value pair. This design reduces computational overhead while maintaining representational capacity, effectively interpolating between MHA ($g = n$) and multi-query qttention ($g = 1$) \citep{shazeer2019fast}.

\subsection{Knocking-Heads Attention}
\label{sec:methods-knocking-heads}
In multi-head attention, each head operates with reduced dimensions, creating a low-rank bottleneck that limits individual head expressiveness \citep{bhojanapalli2020low}. Inspired by talking-heads attention \citep{shazeer2020talking} and head-sharing projections in collaborated attention \citep{cordonnier2020multi}, we propose knocking-heads attention (KHA). After $\mathbf{W}^Q$, $\mathbf{W}^K$, $\mathbf{W}^V$ projections but before individual scaled dot-product attention computations, KHA introduces head-sharing projection matrices (\ie knocking-heads projections) alongside the original head-specific projections that enable head interaction while keeping head specification.

We explore two variants of knocking-heads projections that offer complementary advantages: linear transformations provide inference efficiency through matrix absorption, while MLP transformations offer enhanced expressiveness through non-linearity.

\textbf{KHA-Linear} KHA-Linear applies shared linear transformations to enhance head coordination. For each head $i$, the transformed queries, keys, and values are computed as:
\begin{align}
\tilde{\mathbf{Q}}_i &= \mathbf{Q}_i \mathbf{T}^Q, \quad \mathbf{T}^Q \in \mathbb{R}^{d_k \times d_k}, \\
\tilde{\mathbf{K}}_i &= \mathbf{K}_i \mathbf{T}^K, \quad \mathbf{T}^K \in \mathbb{R}^{d_k \times d_k}, \\
\tilde{\mathbf{V}}_i &= \mathbf{V}_i \mathbf{T}^V, \quad \mathbf{T}^V \in \mathbb{R}^{d_v \times d_v},
\end{align}
where $\mathbf{T}^Q$, $\mathbf{T}^K$, and $\mathbf{T}^V$ are shared transformation matrices across all heads. Crucially, during inference, these transformations can be absorbed into the original projection matrices:
\begin{align}
\mathbf{W}_i^{Q'} &= \mathbf{W}_i^Q \mathbf{T}^Q, \quad \mathbf{W}_i^{K'} = \mathbf{W}_i^K \mathbf{T}^K, \quad \mathbf{W}_i^{V'} = \mathbf{W}_i^V \mathbf{T}^V,
\end{align}
eliminating computational overhead while preserving the benefits of enhanced head coordination.

\textbf{Key Takeway} As we will demonstrate in Section~\ref{sec:attn_variants}, $\mathbf{T}^V$ is the most critical component, as values learn head interactions most effectively. The transformations $\mathbf{T}^Q$ and $\mathbf{T}^K$ are optional, as their removal causes negligible performance degradation.

\textbf{KHA-MLP} To leverage non-linear expressiveness, we introduce an MLP-based transformation that our experiments show outperforms pure linear approaches. Given the parameter overhead of applying MLP to all queries, keys, and values, we focus solely on the most critical values, which maintains the same parameter count as the linear variant while providing superior representational capacity:
\begin{equation}
\tilde{\mathbf{V}}_i = \operatorname{MLP}(\mathbf{V}_i) = 2 \cdot \left(\mathbf{V}_i \mathbf{W}^{\text{up}} \odot \operatorname{Sigmoid}(\mathbf{V}_i \mathbf{W}^{\text{gate}})\right) \mathbf{W}^{\text{down}},
\label{eqn:mlp}
\end{equation}
where $\mathbf{W}^{\text{up}}, \mathbf{W}^{\text{gate}}, \mathbf{W}^{\text{down}} \in \mathbb{R}^{d_v \times d_v}$ are shared across all heads. We use sigmoid-activated MLPs to facilitate zero initialization. Our ablation studies confirm that applying gating alone introduces detrimental effects, while the complete MLP structure enhances model expressiveness.

\subsection{Initialization Strategy}
The effectiveness of KHA relies critically on ``zero" initialization of shared matrices to ensure they approximate identity mappings during early training. Our experiments show that random initialization causes loss to converge to much higher values, potentially making all heads overly similar.

For the KHA-Linear, we apply diagonal initialization to $\mathbf{T}^Q$, $\mathbf{T}^K$, and $\mathbf{T}^V$. For the KHA-MLP, $\mathbf{W}^{\text{up}}$ and $\mathbf{W}^{\text{down}}$ are diagonal-initialized, while $\mathbf{W}^{\text{gate}}$ is zero-initialized. This ensures that $\text{sigmoid}(\mathbf{V}_i \mathbf{W}^{\text{gate}}) \cdot 2 = 1$ initially, which motivates our choice of sigmoid activation in Equation~\ref{eqn:mlp}.

This initialization allows off-diagonal elements to progressively learn non-zero values, enabling the model to first establish head specialization before learning inter-head interactions.

\subsection{Complexity Analysis}
The training FLOPs for multi-head attention can be computed as: 
\begin{align}
\text{FLOP}_{\text{MHA}} = 6Ld^2 + 4nL^2d_k + 2Ld^2 = 8Ld^2 + 4L^2d,
\end{align}
where $6Ld^2$ accounts for query, key, and value projection forward and backward passes, $4nL^2d_k$ represents attention score computation and attention-value multiplication, and $2Ld^2$ corresponds to the output projection matrix $\mathbf{W}^O$. Assuming $d_{\text{ff}} = 3d$, the training FLOPs for the feed-forward network (FFN) with up-gate-down structure can be computed as: $\text{FLOP}_{\text{FFN}} = 6Ld \cdot 3d = 18Ld^2$, where the factor of 6 accounts for forward and backward passes through three linear layers (up, gate, and down projection). Therefore, the total single-layer training FLOPs is:
\begin{align}
\text{FLOP}_{\text{total}} = \text{FLOP}_{\text{MHA}} + \text{FLOP}_{\text{FFN}} = 26Ld^2 + 4L^2d.
\end{align}
For both knocking-heads variants, the additional training FLOPs are identical:
\begin{align}
\text{FLOP}_{\text{KHA}} = \frac{6Ld^2}{n}.
\end{align}
where $n$ is the number of attention heads. For a concrete example with $L = 2048$, $d = 1024$ and $n = 32$, the knocking-heads overhead represents only 0.55\% of the total layer computation and 1.17\% of the original MHA computation, demonstrating the efficiency of our approach.

\section{Experiment}
\subsection{Training Setup}
\label{sec:setup}
We conduct two sets of experiments: main experiments on 1T tokens and exploratory experiments on 100B tokens, both using high-quality multi-domain data.

\paragraph{Architecture} We adopt MoE architectures for most of our experiments, as they consistently outperform dense models under equivalent training FLOPs \citep{dai2024deepseekmoe}. To maintain expert load balancing, we implement an updated loss-free balancing strategy \citep{wang2024auxiliary,su2025moe}. We follow Qwen's design \citep{yang2025qwen3} for attention implementation with QK RMS normalization, and maintain a head dimension of 128 throughout all experiments. For the main experiments, we use a configuration with 128 experts where 8 are activated, yielding 6.1B total parameters with 1.01B active parameters. We employ 32 attention heads with grouped query attention (group size $g=4$) For exploratory experiments, we scale down to 64 experts with 4 activated and vary the hidden dimension to create model variants ranging from A0.44B-2.3B up to A1.6B-14.6B. These experiments evaluate different attention mechanisms including MHA, MQA, GTA, MLA, and GLA, as well as various GQA configurations.

\paragraph{Hyperparameters} All models are trained using Adam optimizer with weight decay 0.1, $\beta_1 = 0.9$, $\beta_2 = 0.95$, and gradient clipping at 1.0. We use FSDP for distributed training. The learning rate schedule includes 5\% warmup steps followed by cosine annealing to 10\% of peak rate at training completion. For main experiments, we use learning rate 4.78e-4, sequence length 8,192, and batch size 4.2M tokens. Exploratory experiments use learning rate 7.5e-4, sequence length 4,096, and batch size 2.1M tokens.

\subsection{Architecture Exploration On Attention}
\label{sec:arch_explore}
We conduct exploratory experiments to evaluate KHA from the following perspectives. First, we examine the impact of different numbers of heads on KHA performance (Section~\ref{sec:different-kv-num}). Second, we explore different positions and methods for applying knocking-heads projections (Section~\ref{sec:kha-variants}). Third, we test the compatibility of KHA with other attention variants (Section~\ref{sec:attn_variants}).
\subsubsection{KHA with Varying Head Number}
\label{sec:different-kv-num}
KHA's core mechanism relies on head-sharing, making it sensitive to the number of attention heads. We evaluate both KHA variants on GQA with varying KV head groups while fixing query heads. All experiments were performed using a 0.8B activated parameter model (6.6B total parameters) trained on 75B tokens. As shown in Table~\ref{tab:num_kv_heads}, the effectiveness of KHA scales with KV head count. Both variants show significant improvements as KV heads increase from 1 to 4, confirming that more heads provide greater opportunities for cross-head information sharing. KHA-MLP show more stable performance across different configurations than the linear-based ones, likely due to the diagonal-initialized MLP introducing beneficial non-linearity beyond head-sharing alone.

\begin{table}[!t]
\caption{Loss of knocking-heads variants with different KV heads number in GQA (32 query heads). All models reported are A0.8-6.6B MoE trained on 75B tokens. $\Delta L$ denotes the loss difference after adopting KHA, where lower values indicate better performance.}
    \centering
    \resizebox{0.7\linewidth}{!}{
    \renewcommand{\arraystretch}{1.3}
    \begin{tabular}{cccccccc}
        \toprule
        \multirow{2}{*}{Model} & \multicolumn{6}{c}{\# KV heads} \\
        \cmidrule(lr){2-7}
        & 1 & 2 & 4 & 8 & 16 & 32 \\
        \midrule
        Baseline & 1.864 & 1.855 & 1.856 & 1.851 & 1.849 & 1.846\\
        KHA-MLP & 1.846 & 1.835 & 1.832 & 1.832 & 1.833 & 1.83\\
        \rowcolor{gray!10} $\Delta L$ & \textbf{-0.017} & \textbf{-0.020} & \textbf{-0.024} & \textbf{-0.019} & \textbf{-0.016} &\textbf{-0.016}\\
        KHA-Linear & 1.857 & 1.845 & 1.838 & 1.841 & 1.832 &1.834\\
        \rowcolor{gray!10} $\Delta L$ & \textbf{-0.007} & \textbf{-0.010} & \textbf{-0.018} & \textbf{-0.010} & \textbf{-0.017} &\textbf{-0.012}\\
        \bottomrule
    \end{tabular}}
    \label{tab:num_kv_heads}
\end{table}

\begin{table}[!t]
\centering
\caption{Loss comparison of knocking-heads variants across different shared projection types (linear, gate, MLP) and positions (query, key, value). All models reported are A0.8-6.6B MoE with GQA ($g=4$) trained on 75B tokens. \textcolor{darkgreen}{Green} indicates the \textbf{best} setting and \textcolor{darkred}{red} indicates the \textbf{worst} setting.}
\label{tab:sharing_config}
\resizebox{0.8\linewidth}{!}{
\begin{tabular}{c*{3}{>{\centering\arraybackslash}p{0.2cm}}c>{\centering\arraybackslash\bfseries}p{1cm}|c*{3}{>{\centering\arraybackslash}p{0.2cm}}c>{\centering\arraybackslash\bfseries}p{1cm}}
\toprule
\multirow{2}{*}{Type} & \multicolumn{3}{c}{Place} & \multirow{2}{*}{Loss} & \multirow{2}{*}{$\Delta L$} & \multirow{2}{*}{Type} & \multicolumn{3}{c}{Place} & \multirow{2}{*}{Loss} & \multirow{2}{*}{$\Delta L$} \\
\cmidrule(lr){2-4} \cmidrule(lr){8-10}
& Q & K & V & &  & & Q & K & V & &  \\
\midrule
- & - & - & - & 1.856 & \cellcolor{gray!10}-- & Gate & - & - & \checkmark & 1.919 & \cellcolor{gray!10}\textcolor{darkred}{+0.063} \\
\cmidrule{1-6} \cmidrule{7-12}
\multirow{4}{*}{Linear} & \checkmark & - & - & 1.849 & \cellcolor{gray!10}-0.007 & \multirow{4}{*}{MLP} & \checkmark & - & - & 1.848 & \cellcolor{gray!10}-0.008 \\
& - & \checkmark & - & 1.848 & \cellcolor{gray!10}-0.008 & & - & \checkmark & - & 1.848 & \cellcolor{gray!10}-0.008 \\
& - & - & \checkmark & 1.839 & \cellcolor{gray!10}-0.017 & & - & - & \checkmark & 1.832 & \cellcolor{gray!10}\textcolor{darkgreen}{-0.024} \\
& \checkmark & \checkmark & \checkmark & 1.838 & \cellcolor{gray!10}-0.018 & & \checkmark & \checkmark & \checkmark & 1.832 & \cellcolor{gray!10}\textcolor{darkgreen}{-0.024} \\
\bottomrule
\end{tabular}}
\label{tab:ablation-knocking-heads-variants}
\end{table}

\subsubsection{Ablation Studies on Different Variants of Knocking-heads}
\label{sec:kha-variants}
We experiment with knocking-heads variants beyond the KHA-Linear and KHA-MLP presented in Section~\ref{sec:methods-knocking-heads}, exploring different architectural choices for inter-head knowledge sharing. The training setup is similar as Section~\ref{sec:different-kv-num} and we fix the group size $g=4$ in GQA based on the results in Table~\ref{tab:num_kv_heads}. Our investigation focused on two key design dimensions: (1) the type of shared transformation matrices (linear layers, gating mechanisms, or MLPs) and (2) their placement within the attention mechanism (query, key, or value projections).

Table~\ref{tab:ablation-knocking-heads-variants} reveals several interesting findings. First, incorporating shared transformations across attention heads consistently improves performance, with benefits observed across all three projection positions (Query, Key, Value) and for both linear and MLP-based transformations. Interestingly, value projections benefit most from the knocking-heads mechanism, suggesting that sharing learned representations in the value space is particularly effective for capturing cross-head dependencies. When comparing transformation types, MLPs demonstrate better performance over linear layers for value projections. This advantage likely stems from non-linear activations in MLPs, but using standalone gating mechanisms as shared transformations proves detrimental to model performance.

\subsubsection{Compatibility with Other Attention Variants}
\label{sec:attn_variants}
KHA can actually adapt to any form of multi-head attention mechanism, but we focus specifically on softmax-based attention here. Except for MLA, which only works with KHA-Linear because they don't explicitly materialize queries/keys/values during inference, other variants are compatible with both KHA-Linear and KHA-MLP. Therefore, for MLA, we use KHA-Linear, while for others, based on the results in Table~\ref{tab:ablation-knocking-heads-variants}, we use KHA-MLP. All experiments were performed using a model with approximately 0.8B activated parameters (around 6.6B total parameters) trained on 100B tokens.

As shown in Table~\ref{tab:attn_variants}, KHA consistently improves all tested variants including GQA, MHA, GTA, and MQA. The results demonstrate the ability of knocking-heads projections to recover performance losses incurred by KV-cache optimizations, allowing models to maintain memory efficiency without sacrificing quality. For example, baseline GQA4(32) underperforms MHA16(16) by 0.012 loss despite using $4\times$ less KV-cache. When both apply knocking-heads projections, GQA4(32) achieves a 0.02 loss reduction, shrinking the gap between GQA4(32) and MHA16(16) to just 0.002. Notably, knocking-heads projections even benefits GTA, which shares non-RoPE features between keys and values, highlighting knocking-heads projection's broad generalizability.
\begin{table}[!t]
\caption{Loss of knocking-heads on various attention variants with different head configurations. All models reported are roughly A0.8-6.6B MoE trained on 100B tokens. $\Delta L$ denotes the loss difference after adopting KHA, where lower values indicate better performance. KHA-MLP is used for all variants except MLA, which uses KHA-Linear.}
\centering
\resizebox{0.9\linewidth}{!}{
    \begin{tabular}{cccccc|ccc}
        \toprule
        \multirow{2}{*}{Model} & Head & \# KV & \# Query & \multirow{2}{*}{Cache} & Activated & \multirow{2}{*}{Baseline} & \multirow{1}{*}{Knocking-} & \multirow{2}{*}{$\Delta L$} \\
        & Dim. & Head & Head & & Params & & Heads & \\
        \midrule
        MLA & 128 & 4 & 8 & 512+64 & 970M & 1.812 & {1.801} & \cellcolor{gray!10}\textbf{-0.011} \\
        GTA & 128 & 4 & 32 & 512+64 & 868M & 1.819 & 1.802 & \cellcolor{gray!10}\textbf{-0.017} \\
        MQA & 128 & 1 & 32 & 256 & 859M & 1.814 & 1.801 & \cellcolor{gray!10}\textbf{-0.013} \\
        GQA & 128 & 8 & 16 & 2048 & 795M & 1.801 & 1.791 & \cellcolor{gray!10}\textbf{-0.010} \\
        GQA & 128 & 4 & 32 & 1024 & 880M & 1.807 & 1.787 & \cellcolor{gray!10}\textbf{-0.020} \\
        MHA & 128 & 16 & 16 & 4096 & 852M & 1.795 & 1.785 & \cellcolor{gray!10}\textbf{-0.010} \\
        \bottomrule
    \end{tabular}}
    \label{tab:attn_variants}
\end{table}

\begin{table}[th!]
\caption{Performance comparison with and without KHA (32 query heads, 4 key/value heads), trained on 1T tokens with 1.01B active and 6.1B total parameters. ``Overall Average" is the average score of all sub-tasks. \textcolor{darkgreen}{Green} values indicate \textbf{improvements}, while \textcolor{darkred}{red}  indicate \textbf{decreases}.}
\small
\label{table:eval}
\centering
\begin{tabular}{clccc}
\toprule
\multicolumn{1}{l}{} & {Metric} & {Baseline} & {Knocking-Heads} & {$\Delta$ Score} \\ \midrule
\multirow{7}{*}{\shortstack{General Knowledge\\Reasoning}} & ARC-challenge & 53.56& 55.25 & \cellcolor{gray!10}\textcolor{darkgreen}{+1.69}\\
& AGIEval & 32.97& 31.33 & \cellcolor{gray!10}\textcolor{darkred}{-1.64}\\
& HellaSwag & 62.44& 62.13 & \cellcolor{gray!10}\textcolor{darkred}{-0.31}\\
& WinoGrande & 60.85& 63.77 & \cellcolor{gray!10}\textcolor{darkgreen}{+2.92}\\
& PIQA & 76.39& 74.86 & \cellcolor{gray!10}\textcolor{darkred}{-1.53}\\ \cmidrule{2-5}
& Average & 57.24 & 57.47 & \cellcolor{gray!10}\textcolor{darkgreen}{+0.23}\\ \midrule
\multirow{7}{*}{\shortstack{Professional\\Knowledge}} & MMLU & 51.22& 51.24 & \cellcolor{gray!10}\textcolor{darkgreen}{+0.02}\\ 
& MMLU-Pro & 23.42& 21.62 & \cellcolor{gray!10}\textcolor{darkred}{-1.80}\\
& CMMLU & 47.52& 47.32 & \cellcolor{gray!10}\textcolor{darkred}{-0.20}\\
& C-Eval & 49.07& 47.02 & \cellcolor{gray!10}\textcolor{darkred}{-2.05}\\
& CommonsenseQA & 59.05& 59.54 & \cellcolor{gray!10}\textcolor{darkgreen}{+0.49}\\
& GPQA & 26.26& 27.27 & \cellcolor{gray!10}\textcolor{darkgreen}{+1.01}\\ \cmidrule{2-5}
& Average & 42.76 & 42.34 & \cellcolor{gray!10}\textcolor{darkred}{-0.42}\\ \midrule
\multirow{3}{*}{\shortstack{Language\\Understanding}} & RACE-middle & 69.08& 73.40 & \cellcolor{gray!10}\textcolor{darkgreen}{+4.32}\\ 
& RACE-high & 61.89& 66.21 & \cellcolor{gray!10}\textcolor{darkgreen}{+4.32}\\ \cmidrule{2-5}
& Average & 65.49 & 69.81 & \cellcolor{gray!10}\textcolor{darkgreen}{+4.32}\\ \midrule
\multirow{6}{*}{Code} & HumanEval-Plus & 35.98& 43.29 & \cellcolor{gray!10}\textcolor{darkgreen}{+7.31}\\
& MBPP & 35.60& 37.60 & \cellcolor{gray!10}\textcolor{darkgreen}{+2.00}\\
& MBPP-Plus & 43.12& 45.50 & \cellcolor{gray!10}\textcolor{darkgreen}{+2.38}\\ \cmidrule{2-5}
& Average & 38.23 & 42.13 & \cellcolor{gray!10}\textcolor{darkgreen}{+3.90}\\ \midrule
\multirow{3}{*}{Math} & GSM8K & 46.17& 47.16 & \cellcolor{gray!10}\textcolor{darkgreen}{+0.99}\\
& MATH & 32.66& 33.44 & \cellcolor{gray!10}\textcolor{darkgreen}{+0.78}\\
& CMATH & 66.30& 69.40 & \cellcolor{gray!10}\textcolor{darkgreen}{+3.10}\\ \cmidrule{2-5}
& Average & 48.38 & 50.00 & \cellcolor{gray!10}\textcolor{darkgreen}{+1.62}\\ \midrule
\multicolumn{2}{c}{Overall Average} & 49.13 & 50.39 & \cellcolor{gray!10}\textcolor{darkgreen}{+1.26}\\ \bottomrule
\end{tabular}
\label{tab:downstream}
\end{table}
\subsection{Large-scale Pretraining On 1T Tokens}
\label{sec:1t_exp}
To validate KHA's effectiveness in large-scale pre-training scenarios, we conducted controlled experiments on a 6.1B-parameter MoE model with 1.01B activated parameters, training on 1T tokens. Based on results in Section~\ref{sec:arch_explore}, we chose a baseline configuration with GQA ($g=4$) and 32 query heads to evaluate adding KHA-MLP's impact. As shown in Fig.~\ref{fig:main}(right), the baseline model exhibited numerous training spikes during the first half of training, whereas applying KHA significantly reduced spike frequency. Furthermore, once loss stabilized in the latter half of training, the model with KHA consistently achieved \textcolor{darkgreen}{\textbf{0.015}} lower loss at equivalent training steps.

We comprehensively evaluated the final trained models on the downstream tasks mentioned in Appendix~\ref{sec:benchmark}. The results in Table~\ref{tab:downstream} demonstrate that while performance on general knowledge and professional knowledge tasks remained comparable between models with and without KHA, significant improvements were observed in language understanding, code, and math tasks, with average score increases of \textcolor{darkgreen}{\textbf{4.32}}, \textcolor{darkgreen}{\textbf{3.9}}, and \textcolor{darkgreen}{\textbf{1.62}} points, respectively. The overall average improvement across all tasks was \textcolor{darkgreen}{\textbf{1.26}} points. In summary, KHA has proven effective in large-scale training, not only substantially reducing training loss spikes but also delivering superior model performance.

\begin{table}[htbp]
\caption{Performance of KHA on different architectures with varying model sizes. All models are trained on 100B tokens and adopt 4 key/value heads. The best results in each row are shown in \textbf{bold}.}
\centering
\resizebox{0.95\linewidth}{!}{
\begin{tabular}{cccccc|cccc}
\toprule
\multirow{2}{*}{Type} & Activated & Total & \multirow{2}{*}{Layers} & Hidden & FFN & \multirow{2}{*}{GQA} & \multirow{1}{*}{KHA-} & \multirow{1}{*}{KHA-} \\
& Params & Params & & Size & Size & & MLP & Linear\\
\midrule
\multirow{3}{*}{MoE} 
& 0.44B & 2.3B & 12 & 1152 & 768×4 & 1.896 & \textbf{1.881$_{-0.015}$} & 1.882$_{-0.014}$ \\
& 0.8B & 6.6B & 18 & 1536 & 1152×4 & 1.807 & \textbf{1.787$_{-0.020}$} & 1.791$_{-0.016}$ \\
& 1.6B & 14.6B & 23 & 2048 & 1536×4 & 1.762 & \textbf{1.737$_{-0.025}$} & 1.742$_{-0.020}$ \\
\midrule
\multirow{3}{*}{Dense} 
& 0.61B & 0.61B & 24 & 1024 & 3072 & 2.017 & 2.013$_{-0.004}$ & \textbf{2.007$_{-0.014}$} \\
& 1.68B & 1.68B & 24 & 2048 & 6144 & 1.892 & 1.884$_{-0.008}$ & \textbf{1.872$_{-0.020}$} \\
& 3.94B & 3.94B & 36 & 2560 & 9728 & 1.815 & 1.811$_{-0.004}$ & \textbf{1.807$_{-0.008}$} \\
\bottomrule
\end{tabular}
}
\label{tab:model_size}
\end{table}
\subsection{Scalability Compared with Transformer}
\label{sec:scalability}
As shown in Table~\ref{tab:model_size}, we evaluate KHA across different model architectures and scales, training MoE models (2.3B-14.6B parameters) and dense models (0.61B-3.94B parameters) on 100B tokens. All models employ GQA with 4 KV heads, 32 query heads, and 128 head dimensions. Knocking-heads delivers substantial improvements for MoE architectures: while scaling baseline MoE from 6.6B to 14.6B parameters reduces loss by 0.045, knocking-heads attention achieves a loss reduction of 0.02 with negligible training overhead. The benefits become more pronounced at larger MoE scales. Additionally, we find that KHA-Linear provides greater improvements on dense models.
\subsection{Visualization}
\subsubsection{Training Stablity}
\begin{figure}[!t]
  \centering
  \begin{subfigure}{0.24\linewidth}
    \centering
    \includegraphics[width=\linewidth]{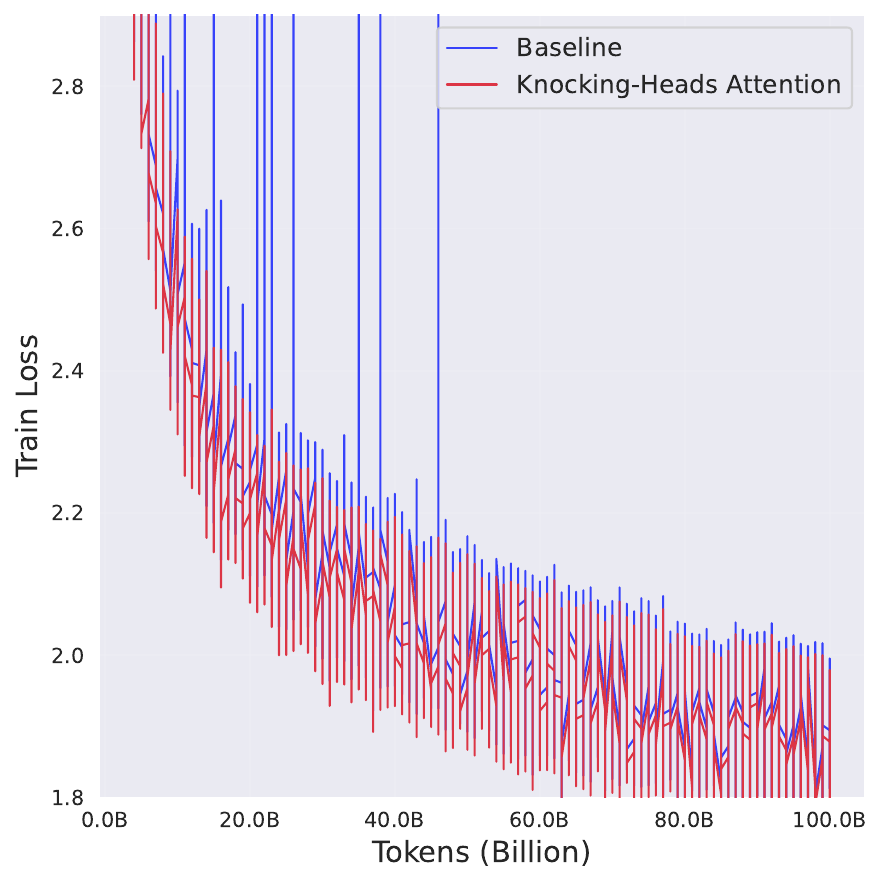}
    \subcaption{A0.44b-2.3B}
  \end{subfigure}
  \begin{subfigure}{0.24\linewidth}
    \centering
    \includegraphics[width=\linewidth]{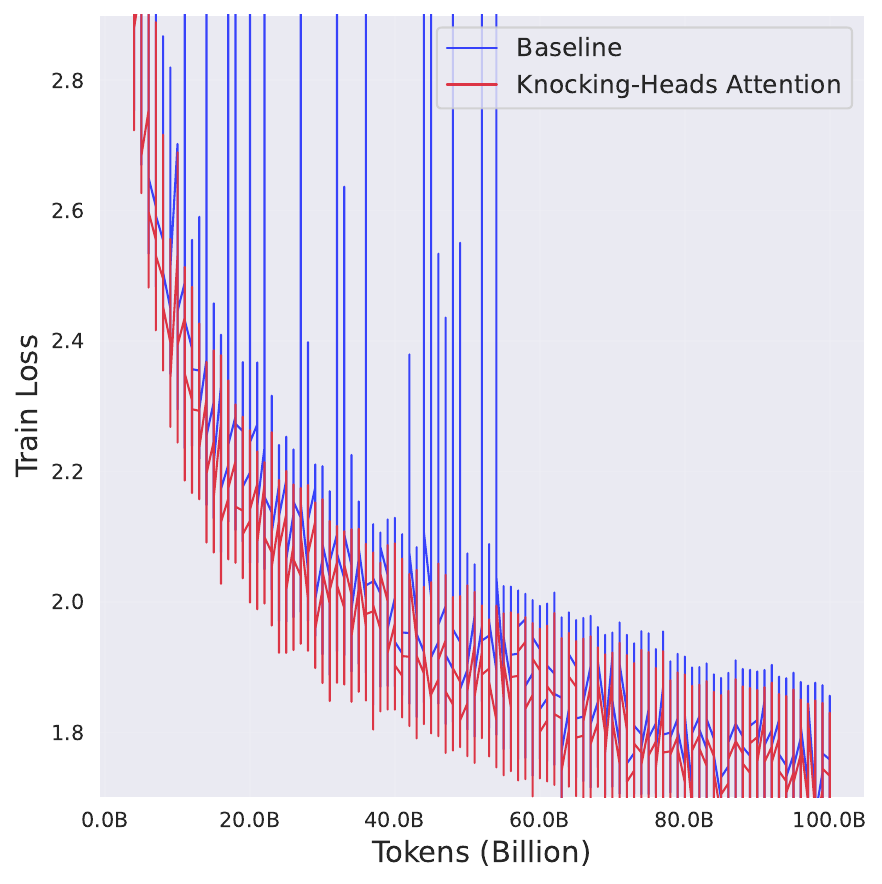}
    \subcaption{A1.6b-14.6B}
  \end{subfigure}
  \vrule
  \begin{subfigure}{0.24\linewidth}
    \centering
    \includegraphics[width=\linewidth]{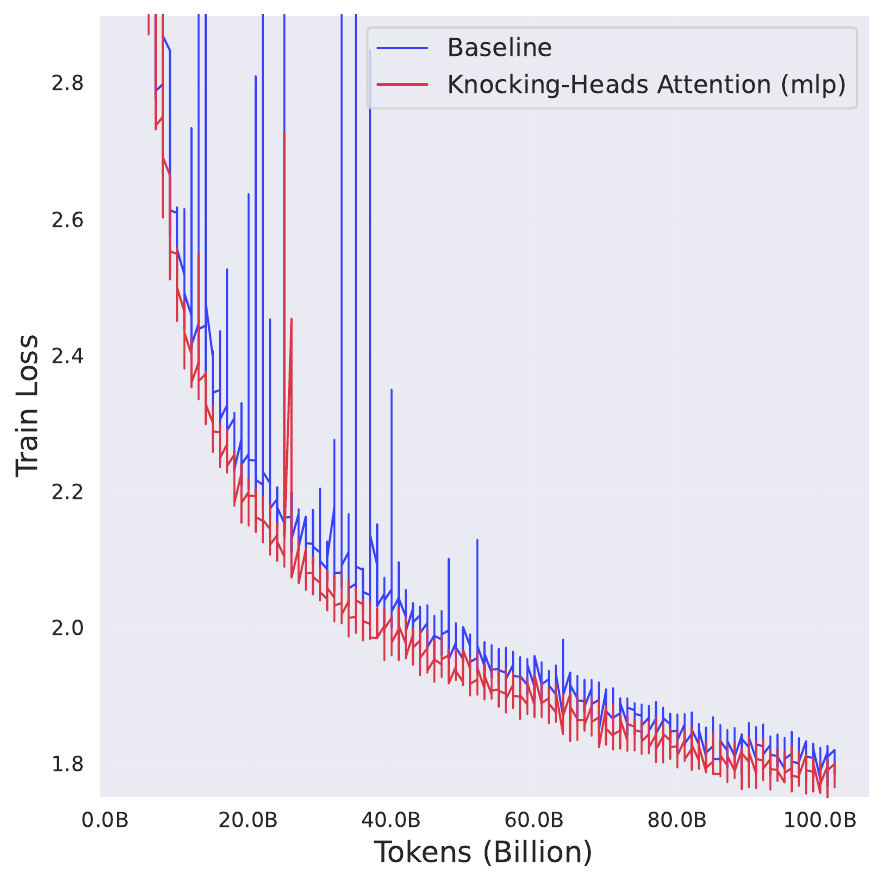}
    \subcaption{A0.8b-6.6B(MLP)}
  \end{subfigure}
  \begin{subfigure}{0.24\linewidth}
    \centering
    \includegraphics[width=\linewidth]{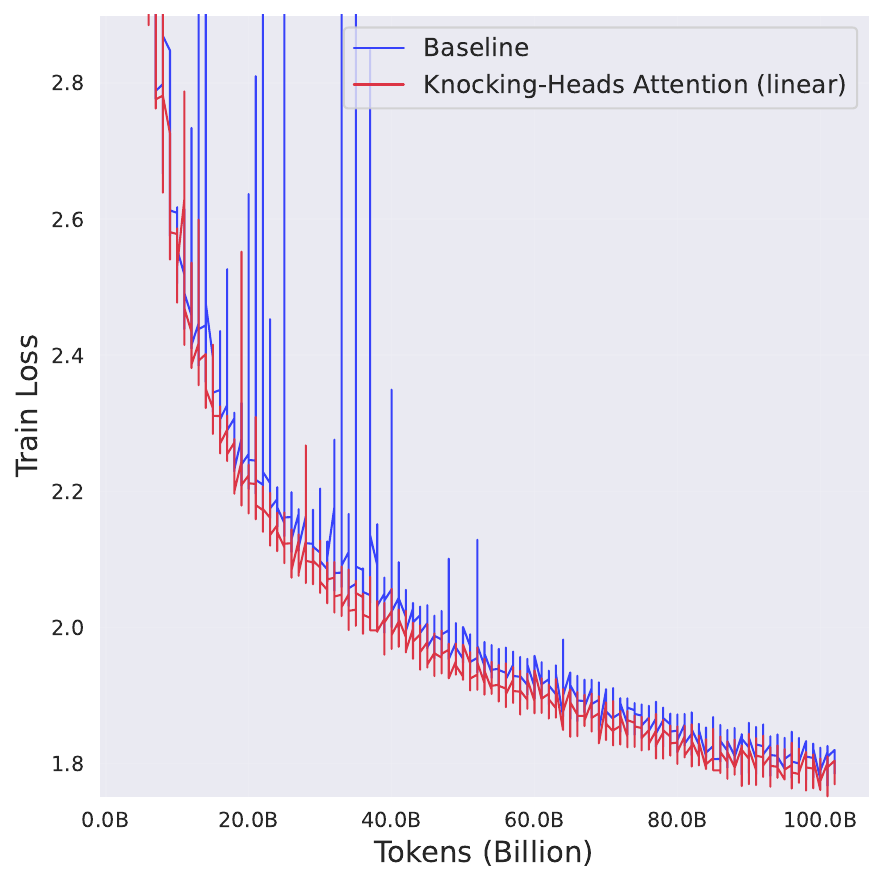}
    \subcaption{A0.8b-6.6B(Linear)}
  \end{subfigure}
\caption{Training loss curves before and after applying knocking-heads across different model sizes, and the loss curves in (c) and (d) are smoothed for better visualization.}
  \label{fig:loss_curves}
\end{figure}

We present the loss curves for selected experiments from Section~\ref{sec:1t_exp} and Section~\ref{sec:scalability} in Fig.\ref{fig:main} and Fig.\ref{fig:loss_curves}, respectively. Loss spikes occur more frequently during the first half of training, with the 1.6B activated MoE model exhibiting significantly higher spike frequency than the 0.4B activated mode. Across all model scales, KHA effectively mitigates loss spikes and improves training stability. For the 0.8B activated MoE model, we compare loss curves of both KHA-Linear and KHA-MLP against the baseline. Both variants successfully reduce loss spikes, confirming that the head-sharing mechanism itself suppresses training instability. We hypothesize this stems from the head-sharing mechanism acting as implicit regularization.
\subsubsection{Learnt Weight of Shared Matrix}
\begin{figure}[!h]
  \centering
  \begin{subfigure}{0.23\linewidth}
    \centering
    \includegraphics[width=\linewidth, trim={2cm 3.5cm 9cm 3cm}, clip]{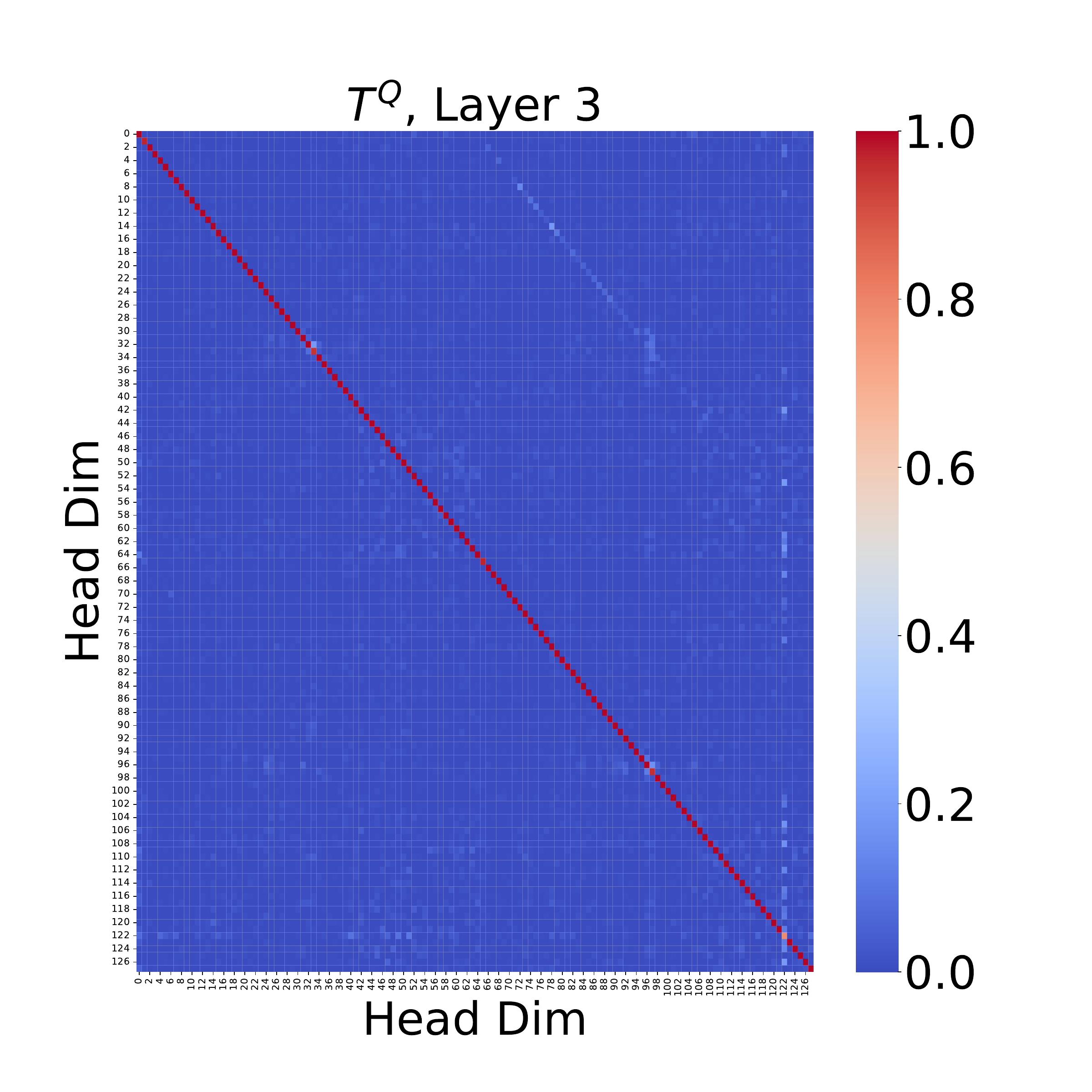}
  \end{subfigure}
  \begin{subfigure}{0.215\linewidth}
    \centering
    \includegraphics[width=\linewidth, trim={3.8cm 3.5cm 9cm 3cm}, clip]{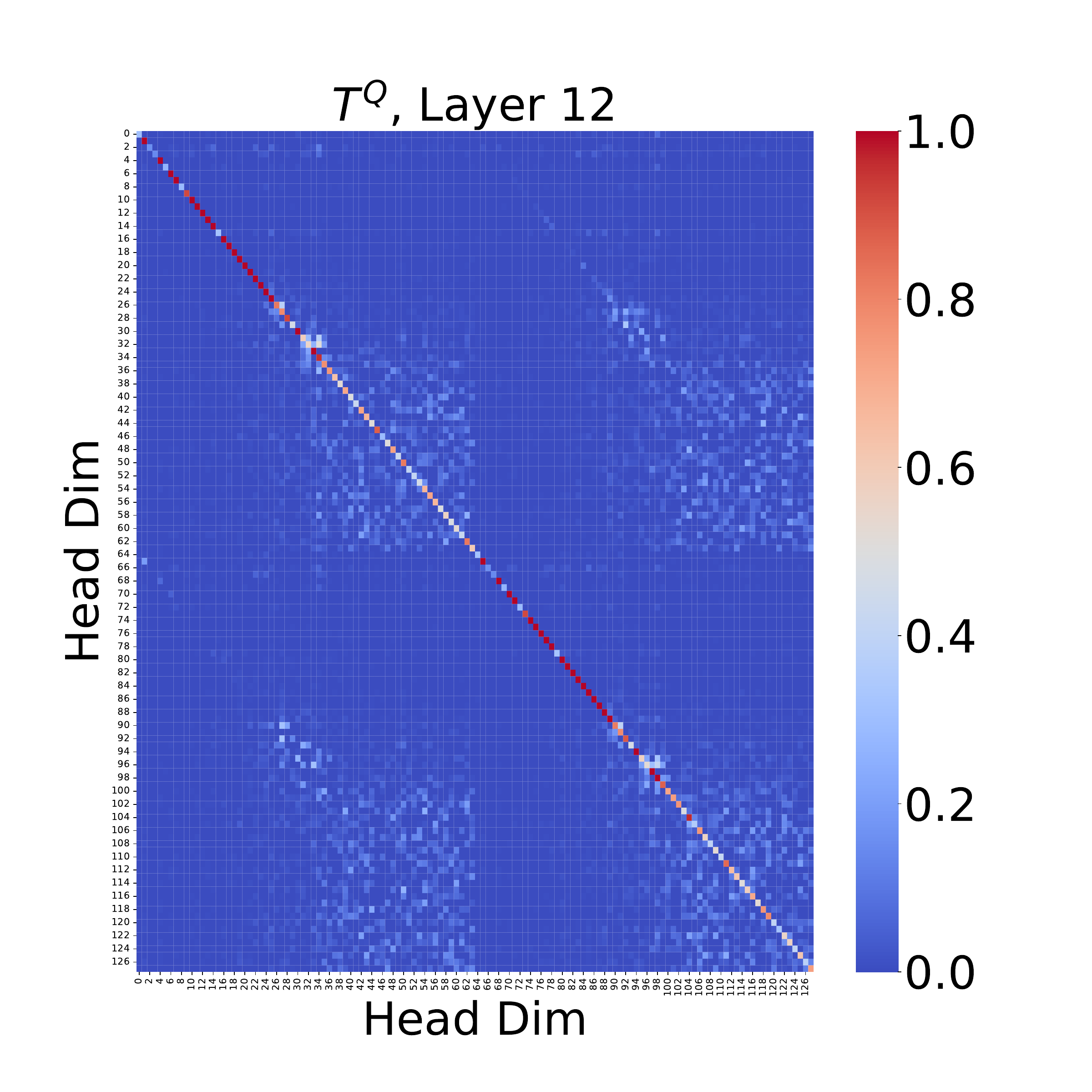}
  \end{subfigure}
  \begin{subfigure}{0.215\linewidth}
    \centering
    \includegraphics[width=\linewidth, trim={3.8cm 3.5cm 9cm 3cm}, clip]{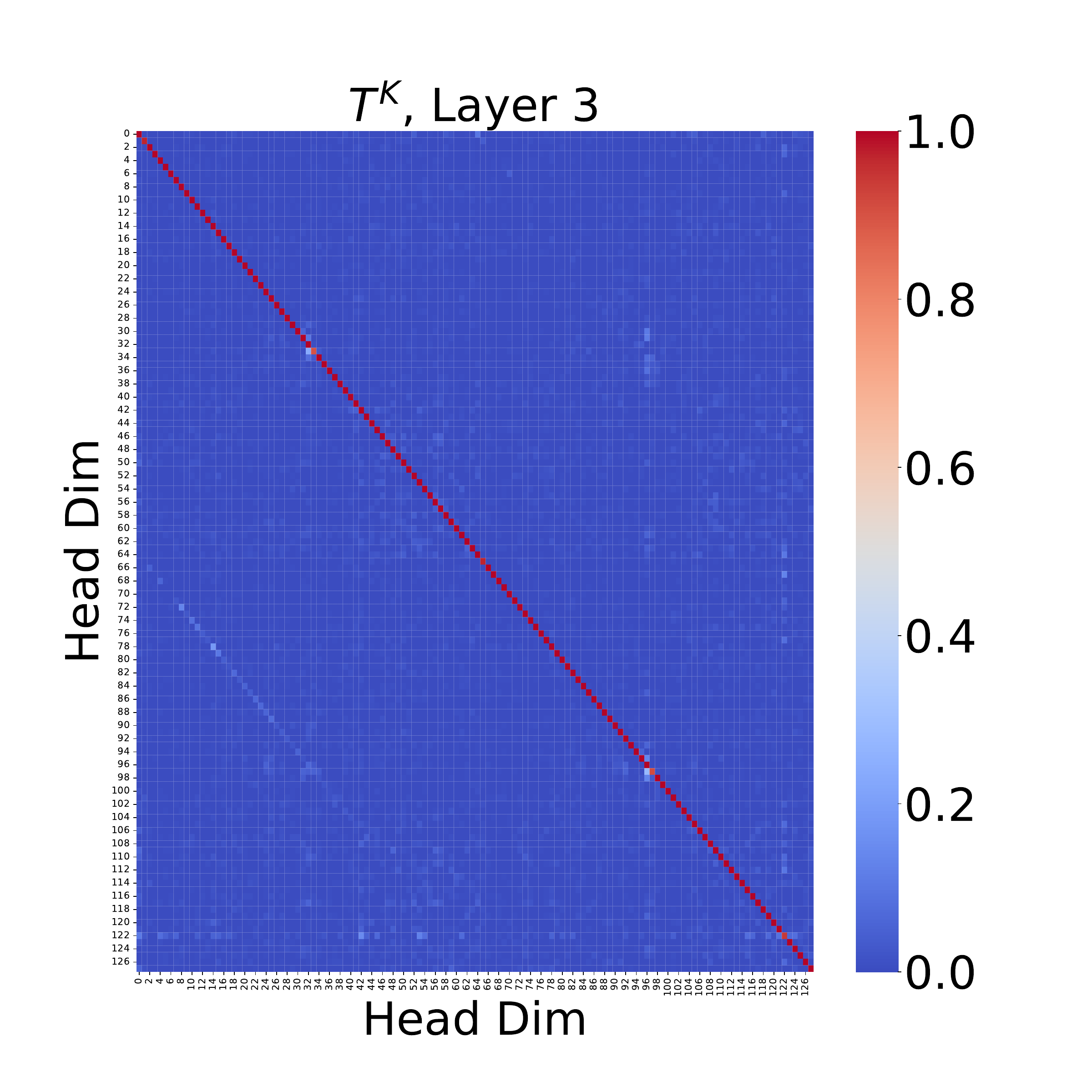}
  \end{subfigure}
  \begin{subfigure}{0.26\linewidth}
    \centering
    \includegraphics[width=\linewidth, trim={3.8cm 3.5cm 3cm 3cm}, clip]{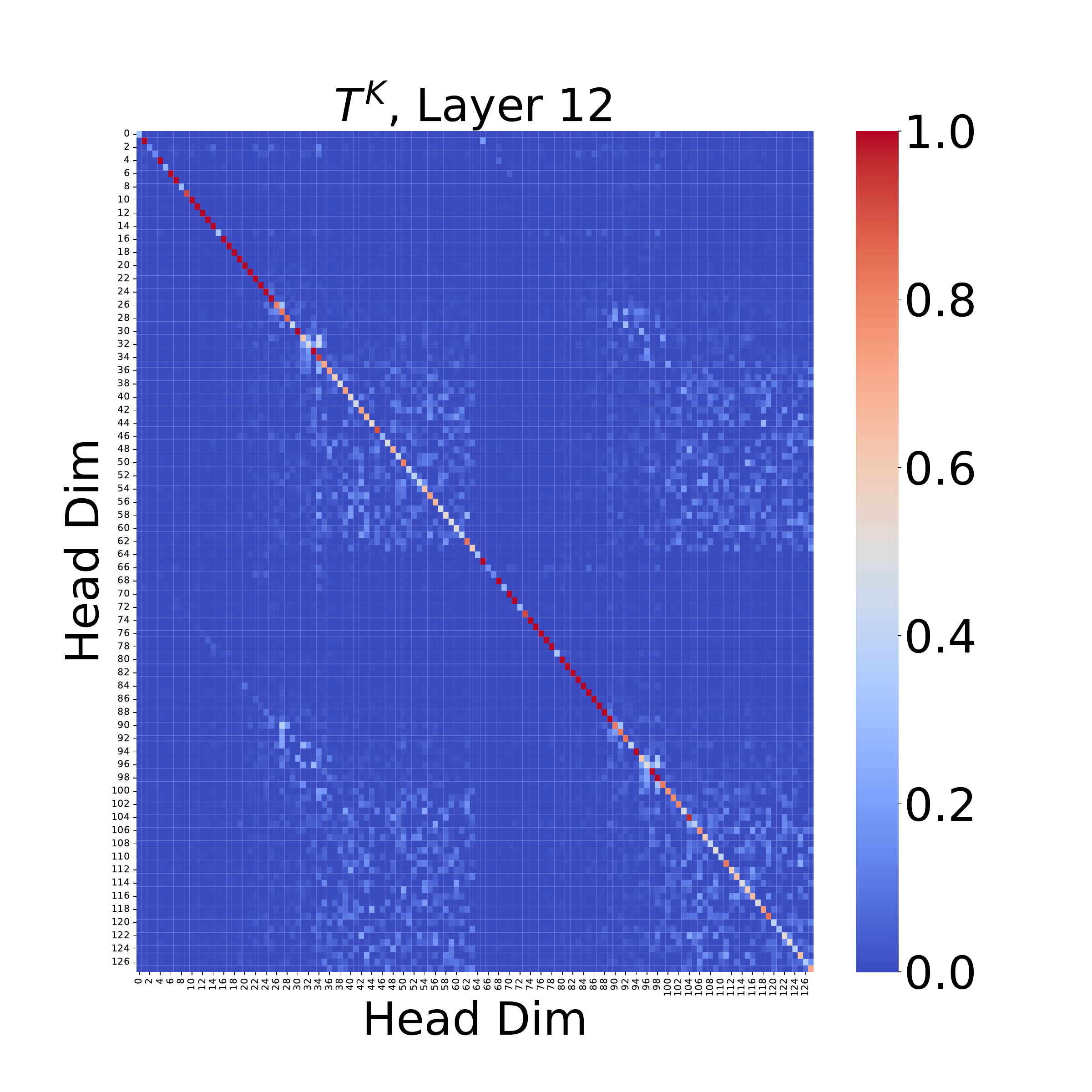}
  \end{subfigure}
  \vspace{2mm}
    \\
  \begin{subfigure}{0.23\linewidth}
    \centering
    \includegraphics[width=\linewidth, trim={2cm 2cm 9cm 3cm}, clip]{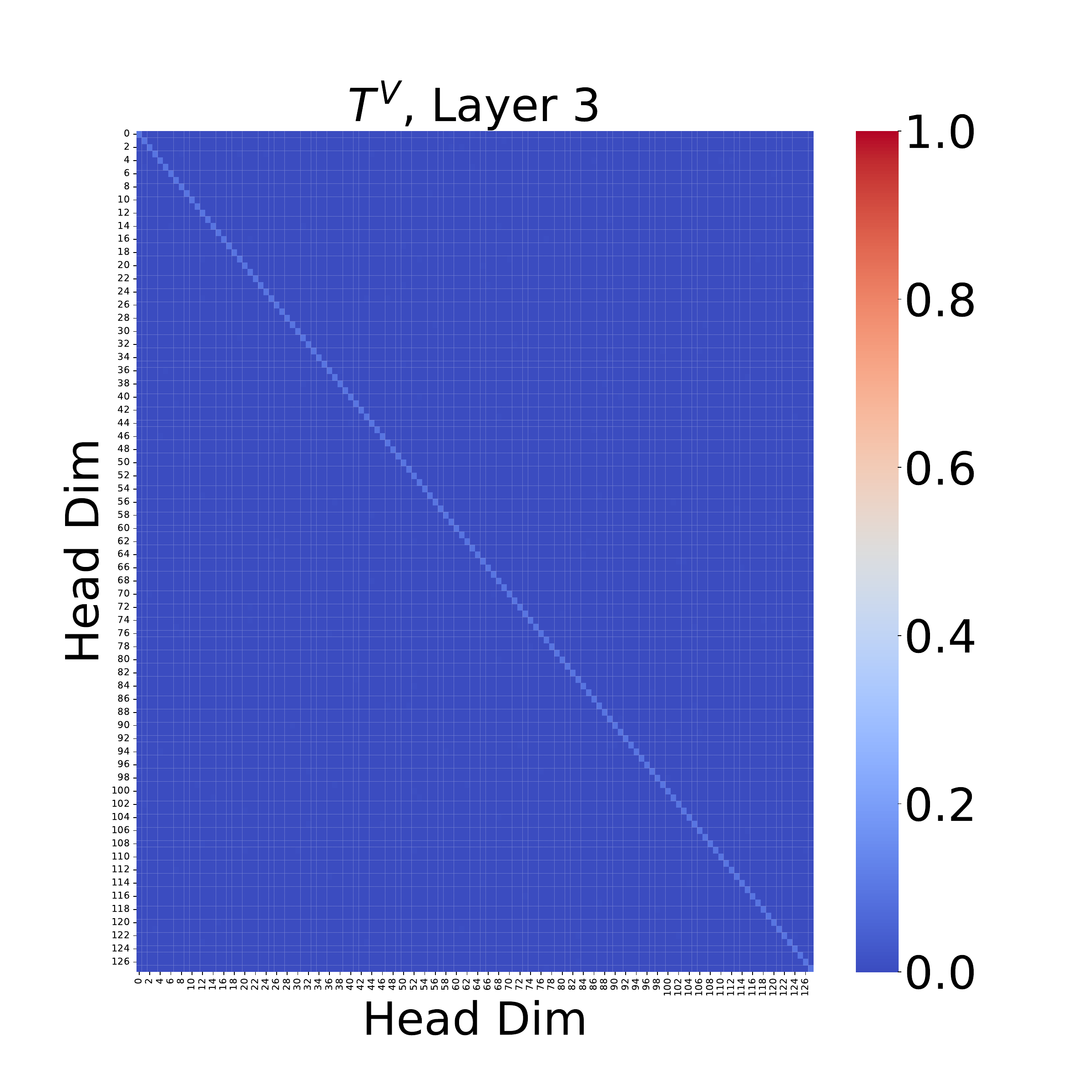}
  \end{subfigure}
  \begin{subfigure}{0.215\linewidth}
    \centering
    \includegraphics[width=\linewidth, trim={3.8cm 2cm 9cm 3cm}, clip]{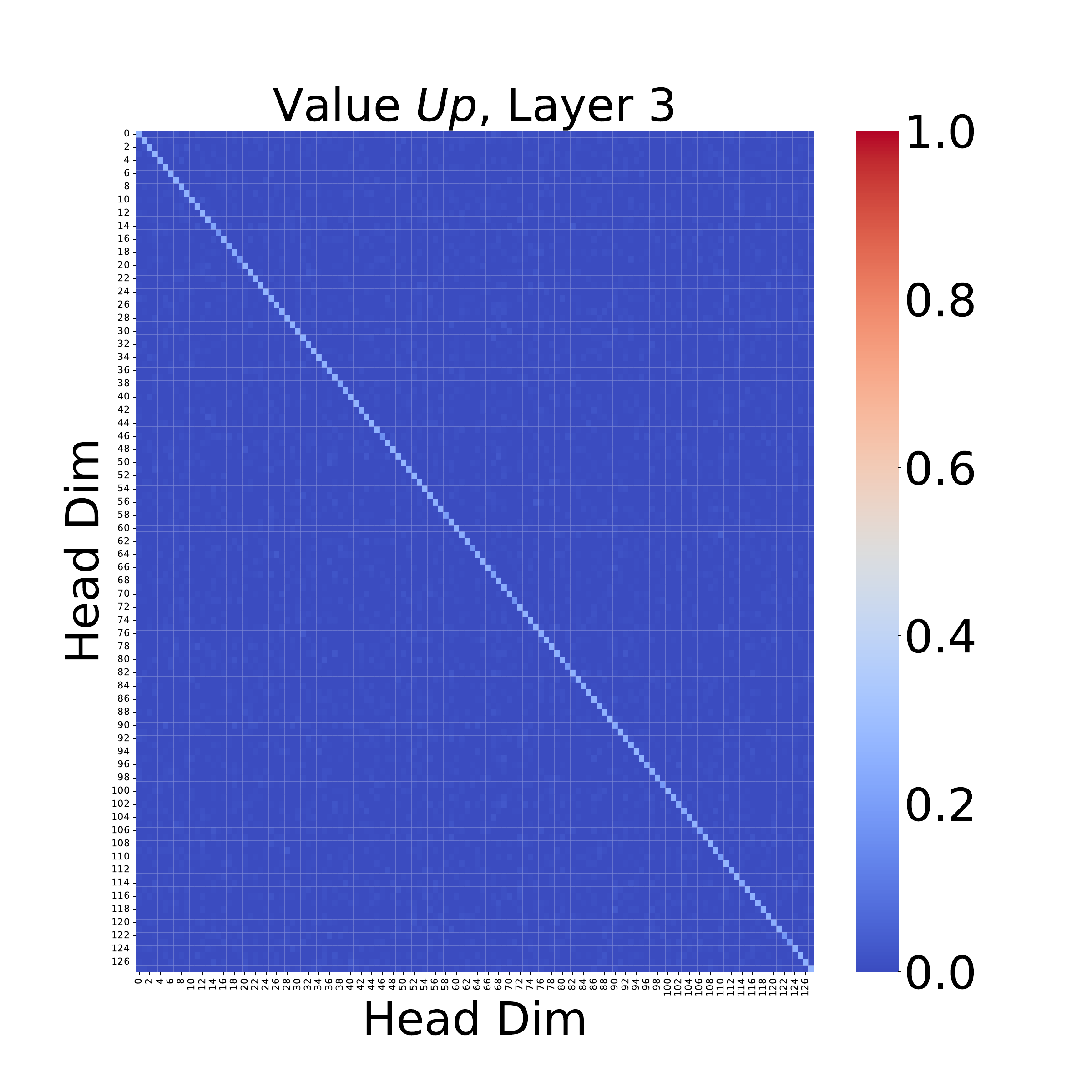}
  \end{subfigure}
  \begin{subfigure}{0.25\linewidth}
    \centering
    \includegraphics[width=\linewidth, trim={3.8cm 2cm 4cm 3cm}, clip]{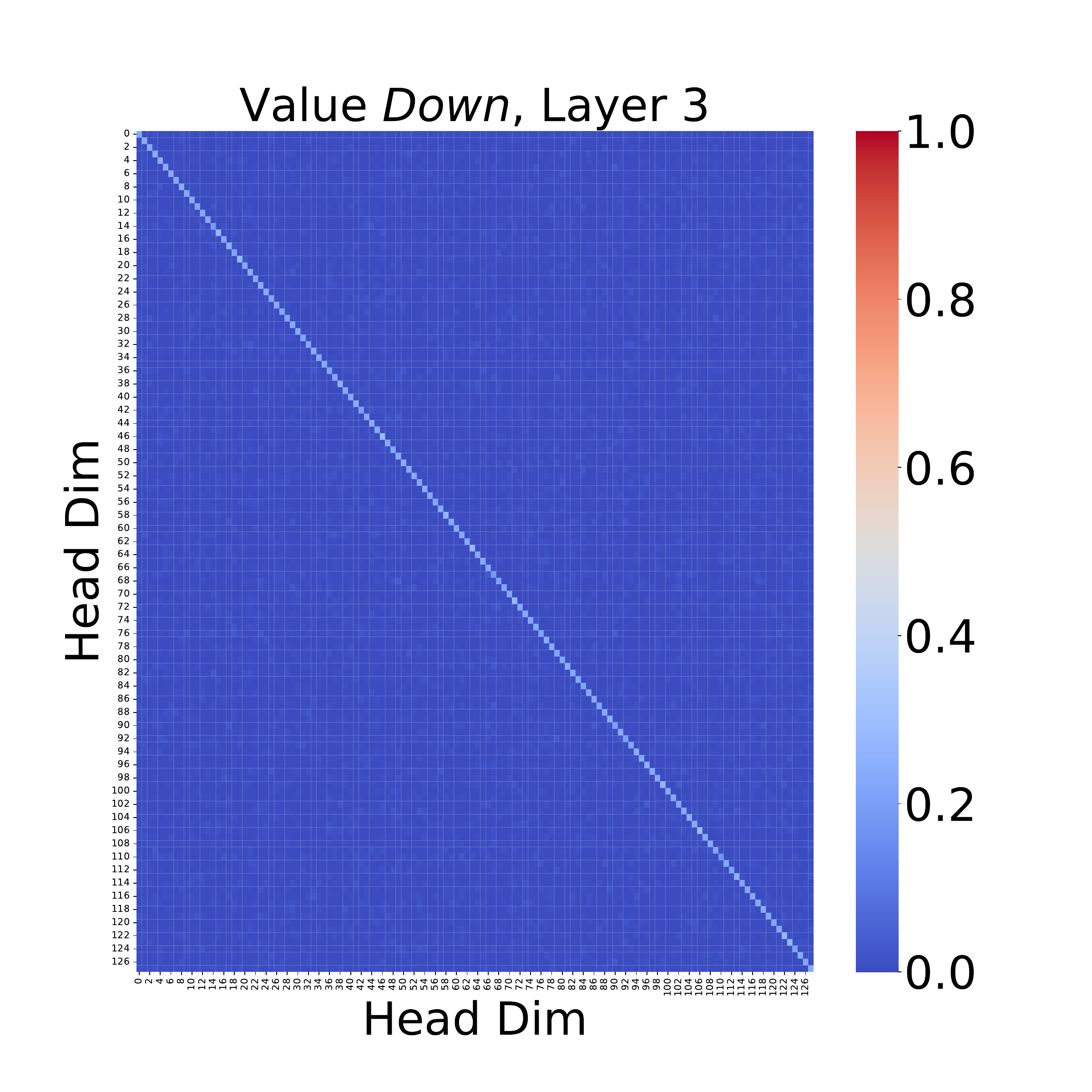}
  \end{subfigure}
  \begin{subfigure}{0.255\linewidth}
    \centering
    \includegraphics[width=\linewidth, trim={3.8cm 2cm 3cm 2.5cm}, clip]{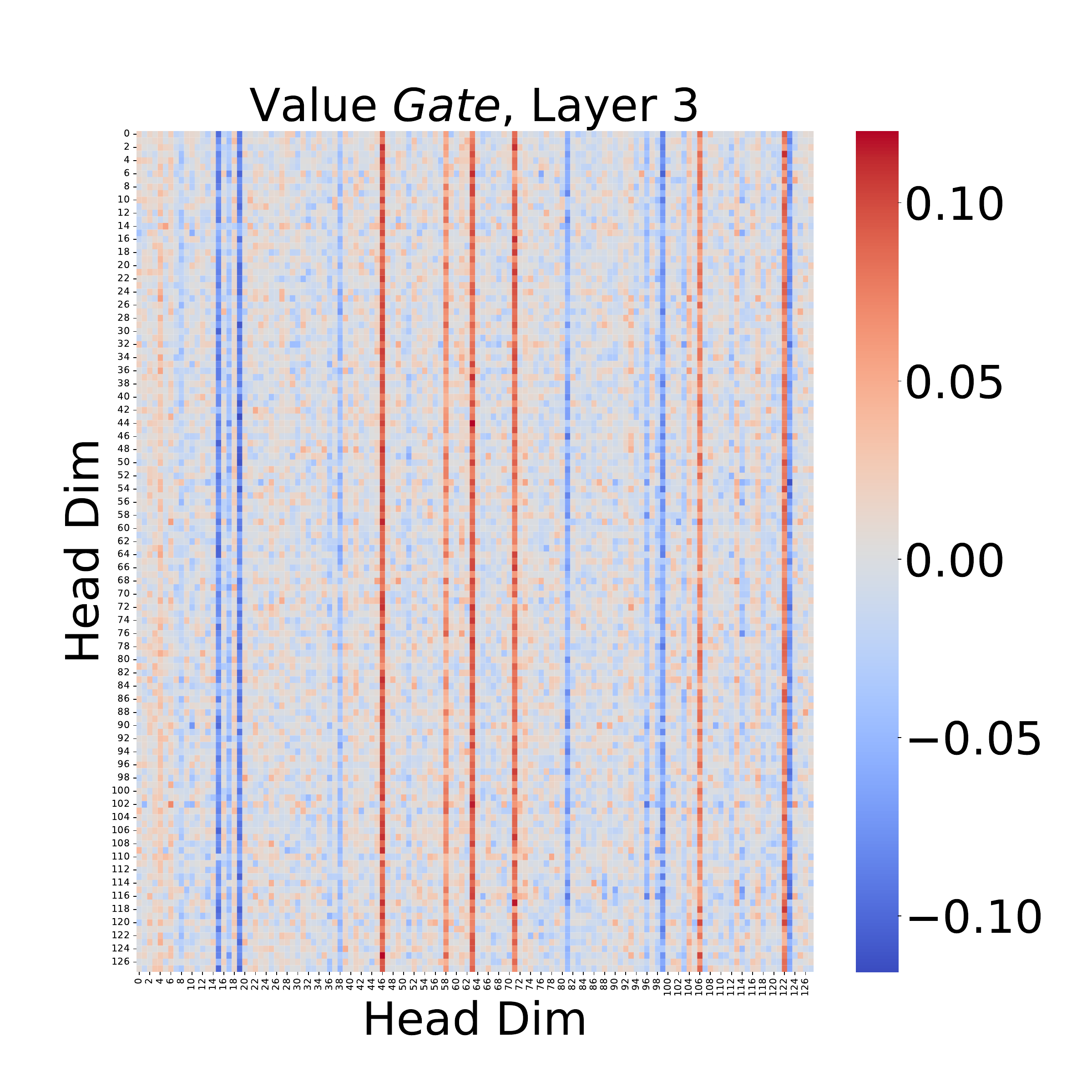}
  \end{subfigure}

    \caption{Visualization of learned knocking-heads projection weights across different layers and types. We apply 0-1 clipping to all knocking-heads projection weights except $W^{gate}$, including $T^K$, $T^Q$, $T^V$, $W^{up}$, and $W^{down}$, for comparative analysis.}
  \label{fig:vis_projections}
\end{figure}

We visualize the learned knocking-heads projection parameters in Fig.~\ref{fig:vis_projections}. $T^Q$ and $T^K$, both applied before QK normalization, show similar block-structured patterns: some feature dimensions exhibit low diagonal values indicating inter-head interaction learning, while others retain high diagonal values for head-specific information. Some layers even exhibit minimal inter-head interaction, highlighting the adaptive nature of our method. Notably, $T^V$ displays a distinct pattern with consistently low diagonal values across layers, indicating aggressive head-sharing. This may explain why value projections yield greater improvements in Table.~\ref{tab:sharing_config}. When using MLPs, the gate matrices also exhibit clear structural patterns.

\section{Conclusion}
In this work, we introduced knocking-heads attention, a simple enhancement to MHA that enables cross-head communication through shared, diagonally-initialized transformation matrices. Our method addresses the limitation that attention heads operate in isolation while adding less than 1\% computational overhead. Through experiments on 1T tokens using 6.1B parameter MoE models, we demonstrate that Knocking-Heads Attention achieves superior performance and significantly improves training stability compared to standard MHA. The diagonal initialization proves crucial for balancing head specialization with cross-head collaboration. As a drop-in replacement for standard MHA, our approach offers a practical enhancement for transformer architectures.

% Reference
\clearpage 
\bibliography{KHA}

\begin{thebibliography}{39}
\providecommand{\natexlab}[1]{#1}
\providecommand{\url}[1]{\texttt{#1}}
\expandafter\ifx\csname urlstyle\endcsname\relax
  \providecommand{\doi}[1]{doi: #1}\else
  \providecommand{\doi}{doi: \begingroup \urlstyle{rm}\Url}\fi

\bibitem[Ainslie et~al.(2023)Ainslie, Lee-Thorp, De~Jong, Zemlyanskiy, Lebr{\'o}n, and Sanghai]{ainslie2023gqa}
Joshua Ainslie, James Lee-Thorp, Michiel De~Jong, Yury Zemlyanskiy, Federico Lebr{\'o}n, and Sumit Sanghai.
\newblock Gqa: Training generalized multi-query transformer models from multi-head checkpoints.
\newblock \emph{arXiv preprint arXiv:2305.13245}, 2023.

\bibitem[Bhakthavatsalam et~al.(2021)Bhakthavatsalam, Khashabi, Khot, Mishra, Richardson, Sabharwal, Schoenick, Tafjord, and Clark]{arc}
Sumithra Bhakthavatsalam, Daniel Khashabi, Tushar Khot, Bhavana~Dalvi Mishra, Kyle Richardson, Ashish Sabharwal, Carissa Schoenick, Oyvind Tafjord, and Peter Clark.
\newblock Think you have solved direct-answer question answering? try arc-da, the direct-answer {AI2} reasoning challenge.
\newblock \emph{CoRR}, abs/2102.03315, 2021.
\newblock URL \url{https://arxiv.org/abs/2102.03315}.

\bibitem[Bhojanapalli et~al.(2020)Bhojanapalli, Yun, Rawat, Reddi, and Kumar]{bhojanapalli2020low}
Srinadh Bhojanapalli, Chulhee Yun, Ankit~Singh Rawat, Sashank Reddi, and Sanjiv Kumar.
\newblock Low-rank bottleneck in multi-head attention models.
\newblock In \emph{International conference on machine learning}, pp.\  864--873. PMLR, 2020.

\bibitem[Bisk et~al.(2020)Bisk, Zellers, Bras, Gao, and Choi]{piqa}
Yonatan Bisk, Rowan Zellers, Ronan~Le Bras, Jianfeng Gao, and Yejin Choi.
\newblock {PIQA:} reasoning about physical commonsense in natural language.
\newblock In \emph{The Thirty-Fourth {AAAI} Conference on Artificial Intelligence, {AAAI} 2020, The Thirty-Second Innovative Applications of Artificial Intelligence Conference, {IAAI} 2020, The Tenth {AAAI} Symposium on Educational Advances in Artificial Intelligence, {EAAI} 2020, New York, NY, USA, February 7-12, 2020}, pp.\  7432--7439. {AAAI} Press, 2020.
\newblock \doi{10.1609/AAAI.V34I05.6239}.
\newblock URL \url{https://doi.org/10.1609/aaai.v34i05.6239}.

\bibitem[Chowdhery et~al.(2023)Chowdhery, Narang, Devlin, Bosma, Mishra, Roberts, Barham, Chung, Sutton, Gehrmann, et~al.]{chowdhery2023palm}
Aakanksha Chowdhery, Sharan Narang, Jacob Devlin, Maarten Bosma, Gaurav Mishra, Adam Roberts, Paul Barham, Hyung~Won Chung, Charles Sutton, Sebastian Gehrmann, et~al.
\newblock Palm: Scaling language modeling with pathways.
\newblock \emph{Journal of Machine Learning Research}, 24\penalty0 (240):\penalty0 1--113, 2023.

\bibitem[Cobbe et~al.(2021)Cobbe, Kosaraju, Bavarian, Chen, Jun, Kaiser, Plappert, Tworek, Hilton, Nakano, Hesse, and Schulman]{gsm8k}
Karl Cobbe, Vineet Kosaraju, Mohammad Bavarian, Mark Chen, Heewoo Jun, Lukasz Kaiser, Matthias Plappert, Jerry Tworek, Jacob Hilton, Reiichiro Nakano, Christopher Hesse, and John Schulman.
\newblock Training verifiers to solve math word problems.
\newblock \emph{CoRR}, abs/2110.14168, 2021.
\newblock URL \url{https://arxiv.org/abs/2110.14168}.

\bibitem[Cordonnier et~al.(2020)Cordonnier, Loukas, and Jaggi]{cordonnier2020multi}
Jean-Baptiste Cordonnier, Andreas Loukas, and Martin Jaggi.
\newblock Multi-head attention: Collaborate instead of concatenate.
\newblock \emph{arXiv preprint arXiv:2006.16362}, 2020.

\bibitem[Dai et~al.(2024)Dai, Deng, Zhao, Xu, Gao, Chen, Li, Zeng, Yu, Wu, et~al.]{dai2024deepseekmoe}
Damai Dai, Chengqi Deng, Chenggang Zhao, RX~Xu, Huazuo Gao, Deli Chen, Jiashi Li, Wangding Zeng, Xingkai Yu, Yu~Wu, et~al.
\newblock Deepseekmoe: Towards ultimate expert specialization in mixture-of-experts language models.
\newblock \emph{arXiv preprint arXiv:2401.06066}, 2024.

\bibitem[Dao et~al.(2022)Dao, Fu, Ermon, Rudra, and R{\'e}]{dao2022flashattention}
Tri Dao, Dan Fu, Stefano Ermon, Atri Rudra, and Christopher R{\'e}.
\newblock Flashattention: Fast and memory-efficient exact attention with io-awareness.
\newblock \emph{Advances in neural information processing systems}, 35:\penalty0 16344--16359, 2022.

\bibitem[Hendrycks et~al.(2021{\natexlab{a}})Hendrycks, Burns, Basart, Zou, Mazeika, Song, and Steinhardt]{mmlu}
Dan Hendrycks, Collin Burns, Steven Basart, Andy Zou, Mantas Mazeika, Dawn Song, and Jacob Steinhardt.
\newblock Measuring massive multitask language understanding.
\newblock In \emph{9th International Conference on Learning Representations, {ICLR} 2021, Virtual Event, Austria, May 3-7, 2021}. OpenReview.net, 2021{\natexlab{a}}.
\newblock URL \url{https://openreview.net/forum?id=d7KBjmI3GmQ}.

\bibitem[Hendrycks et~al.(2021{\natexlab{b}})Hendrycks, Burns, Kadavath, Arora, Basart, Tang, Song, and Steinhardt]{math}
Dan Hendrycks, Collin Burns, Saurav Kadavath, Akul Arora, Steven Basart, Eric Tang, Dawn Song, and Jacob Steinhardt.
\newblock Measuring mathematical problem solving with the {MATH} dataset.
\newblock In Joaquin Vanschoren and Sai{-}Kit Yeung (eds.), \emph{Proceedings of the Neural Information Processing Systems Track on Datasets and Benchmarks 1, NeurIPS Datasets and Benchmarks 2021, December 2021, virtual}, 2021{\natexlab{b}}.
\newblock URL \url{https://datasets-benchmarks-proceedings.neurips.cc/paper/2021/hash/be83ab3ecd0db773eb2dc1b0a17836a1-Abstract-round2.html}.

\bibitem[Huang et~al.(2023)Huang, Bai, Zhu, Zhang, Zhang, Su, Liu, Lv, Zhang, Lei, Fu, Sun, and He]{ceval}
Yuzhen Huang, Yuzhuo Bai, Zhihao Zhu, Junlei Zhang, Jinghan Zhang, Tangjun Su, Junteng Liu, Chuancheng Lv, Yikai Zhang, Jiayi Lei, Yao Fu, Maosong Sun, and Junxian He.
\newblock C-eval: {A} multi-level multi-discipline chinese evaluation suite for foundation models.
\newblock In Alice Oh, Tristan Naumann, Amir Globerson, Kate Saenko, Moritz Hardt, and Sergey Levine (eds.), \emph{Advances in Neural Information Processing Systems 36: Annual Conference on Neural Information Processing Systems 2023, NeurIPS 2023, New Orleans, LA, USA, December 10 - 16, 2023}, 2023.
\newblock URL \url{http://papers.nips.cc/paper\_files/paper/2023/hash/c6ec1844bec96d6d32ae95ae694e23d8-Abstract-Datasets\_and\_Benchmarks.html}.

\bibitem[Jin et~al.(2024)Jin, Zhu, Yuan, and Yan]{jin2024moh}
Peng Jin, Bo~Zhu, Li~Yuan, and Shuicheng Yan.
\newblock Moh: Multi-head attention as mixture-of-head attention.
\newblock \emph{arXiv preprint arXiv:2410.11842}, 2024.

\bibitem[Lai et~al.(2017)Lai, Xie, Liu, Yang, and Hovy]{race}
Guokun Lai, Qizhe Xie, Hanxiao Liu, Yiming Yang, and Eduard~H. Hovy.
\newblock {RACE:} large-scale reading comprehension dataset from examinations.
\newblock In Martha Palmer, Rebecca Hwa, and Sebastian Riedel (eds.), \emph{Proceedings of the 2017 Conference on Empirical Methods in Natural Language Processing, {EMNLP} 2017, Copenhagen, Denmark, September 9-11, 2017}, pp.\  785--794. Association for Computational Linguistics, 2017.
\newblock \doi{10.18653/V1/D17-1082}.
\newblock URL \url{https://doi.org/10.18653/v1/d17-1082}.

\bibitem[Lan et~al.(2020)Lan, Chen, Goodman, Gimpel, Sharma, and Soricut]{lan2019albert}
Zhenzhong Lan, Mingda Chen, Sebastian Goodman, Kevin Gimpel, Piyush Sharma, and Radu Soricut.
\newblock {ALBERT:} {A} lite {BERT} for self-supervised learning of language representations.
\newblock In \emph{8th International Conference on Learning Representations, {ICLR} 2020, Addis Ababa, Ethiopia, April 26-30, 2020}. OpenReview.net, 2020.

\bibitem[LeCun et~al.(2002)LeCun, Bottou, Bengio, and Haffner]{lecun2002gradient}
Yann LeCun, L{\'e}on Bottou, Yoshua Bengio, and Patrick Haffner.
\newblock Gradient-based learning applied to document recognition.
\newblock \emph{Proceedings of the IEEE}, 86\penalty0 (11):\penalty0 2278--2324, 2002.

\bibitem[Li et~al.(2024)Li, Zhang, Koto, Yang, Zhao, Gong, Duan, and Baldwin]{cmmlu}
Haonan Li, Yixuan Zhang, Fajri Koto, Yifei Yang, Hai Zhao, Yeyun Gong, Nan Duan, and Timothy Baldwin.
\newblock {CMMLU:} measuring massive multitask language understanding in chinese.
\newblock In Lun{-}Wei Ku, Andre Martins, and Vivek Srikumar (eds.), \emph{Findings of the Association for Computational Linguistics, {ACL} 2024, Bangkok, Thailand and virtual meeting, August 11-16, 2024}, pp.\  11260--11285. Association for Computational Linguistics, 2024.
\newblock \doi{10.18653/V1/2024.FINDINGS-ACL.671}.
\newblock URL \url{https://doi.org/10.18653/v1/2024.findings-acl.671}.

\bibitem[Liu et~al.(2024)Liu, Feng, Wang, Wang, Liu, Zhao, Dengr, Ruan, Dai, Guo, et~al.]{liu2024deepseek}
Aixin Liu, Bei Feng, Bin Wang, Bingxuan Wang, Bo~Liu, Chenggang Zhao, Chengqi Dengr, Chong Ruan, Damai Dai, Daya Guo, et~al.
\newblock Deepseek-v2: A strong, economical, and efficient mixture-of-experts language model.
\newblock \emph{arXiv preprint arXiv:2405.04434}, 2024.

\bibitem[Liu et~al.(2023)Liu, Xia, Wang, and Zhang]{evalplus}
Jiawei Liu, Chunqiu~Steven Xia, Yuyao Wang, and Lingming Zhang.
\newblock Is your code generated by chat{GPT} really correct? rigorous evaluation of large language models for code generation.
\newblock In \emph{Thirty-seventh Conference on Neural Information Processing Systems}, 2023.
\newblock URL \url{https://openreview.net/forum?id=1qvx610Cu7}.

\bibitem[Qin et~al.(2025)Qin, Shen, and Zhong]{qin2025elucidating}
Zhen Qin, Xuyang Shen, and Yiran Zhong.
\newblock Elucidating the design space of decay in linear attention.
\newblock \emph{arXiv preprint arXiv:2509.05282}, 2025.

\bibitem[Qiu et~al.(2025)Qiu, Wang, Zheng, Huang, Wen, Yang, Men, Yu, Huang, Huang, et~al.]{qiu2025gated}
Zihan Qiu, Zekun Wang, Bo~Zheng, Zeyu Huang, Kaiyue Wen, Songlin Yang, Rui Men, Le~Yu, Fei Huang, Suozhi Huang, et~al.
\newblock Gated attention for large language models: Non-linearity, sparsity, and attention-sink-free.
\newblock \emph{arXiv preprint arXiv:2505.06708}, 2025.

\bibitem[Rein et~al.(2023)Rein, Hou, Stickland, Petty, Pang, Dirani, Michael, and Bowman]{gpqa}
David Rein, Betty~Li Hou, Asa~Cooper Stickland, Jackson Petty, Richard~Yuanzhe Pang, Julien Dirani, Julian Michael, and Samuel~R. Bowman.
\newblock {GPQA:} {A} graduate-level google-proof q{\&}a benchmark.
\newblock \emph{CoRR}, abs/2311.12022, 2023.
\newblock \doi{10.48550/ARXIV.2311.12022}.
\newblock URL \url{https://doi.org/10.48550/arXiv.2311.12022}.

\bibitem[Shazeer(2019)]{shazeer2019fast}
Noam Shazeer.
\newblock Fast transformer decoding: One write-head is all you need.
\newblock \emph{arXiv preprint arXiv:1911.02150}, 2019.

\bibitem[Shazeer et~al.(2020)Shazeer, Lan, Cheng, Ding, and Hou]{shazeer2020talking}
Noam Shazeer, Zhenzhong Lan, Youlong Cheng, Nan Ding, and Le~Hou.
\newblock Talking-heads attention.
\newblock \emph{arXiv preprint arXiv:2003.02436}, 2020.

\bibitem[Su(2025)]{su2025moe}
Jianlin Su.
\newblock Moe travels 3, 3 2025.
\newblock URL \url{https://spaces.ac.cn/archives/10757}.

\bibitem[Takase et~al.(2023)Takase, Kiyono, Kobayashi, and Suzuki]{takase2023spike}
Sho Takase, Shun Kiyono, Sosuke Kobayashi, and Jun Suzuki.
\newblock Spike no more: Stabilizing the pre-training of large language models.
\newblock \emph{arXiv preprint arXiv:2312.16903}, 2023.

\bibitem[Talmor et~al.(2018)Talmor, Herzig, Lourie, and Berant]{talmor2018commonsenseqa}
Alon Talmor, Jonathan Herzig, Nicholas Lourie, and Jonathan Berant.
\newblock Commonsenseqa: A question answering challenge targeting commonsense knowledge.
\newblock \emph{arXiv preprint arXiv:1811.00937}, 2018.

\bibitem[Tao et~al.(2024)Tao, Ventresque, Nallur, and Saber]{mbpp}
Ning Tao, Anthony Ventresque, Vivek Nallur, and Takfarinas Saber.
\newblock Enhancing program synthesis with large language models using many-objective grammar-guided genetic programming.
\newblock \emph{Algorithms}, 17\penalty0 (7):\penalty0 287, 2024.
\newblock \doi{10.3390/A17070287}.
\newblock URL \url{https://doi.org/10.3390/a17070287}.

\bibitem[Team et~al.(2025)Team, Bai, Bao, Chen, Chen, Chen, Chen, Chen, Chen, Chen, et~al.]{team2025kimi}
Kimi Team, Yifan Bai, Yiping Bao, Guanduo Chen, Jiahao Chen, Ningxin Chen, Ruijue Chen, Yanru Chen, Yuankun Chen, Yutian Chen, et~al.
\newblock Kimi k2: Open agentic intelligence.
\newblock \emph{arXiv preprint arXiv:2507.20534}, 2025.

\bibitem[Vaswani et~al.(2017)Vaswani, Shazeer, Parmar, Uszkoreit, Jones, Gomez, Kaiser, and Polosukhin]{vaswani2017attention}
Ashish Vaswani, Noam Shazeer, Niki Parmar, Jakob Uszkoreit, Llion Jones, Aidan~N Gomez, {\L}ukasz Kaiser, and Illia Polosukhin.
\newblock Attention is all you need.
\newblock \emph{Advances in neural information processing systems}, 30, 2017.

\bibitem[Wang et~al.(2024{\natexlab{a}})Wang, Gao, Zhao, Sun, and Dai]{wang2024auxiliary}
Lean Wang, Huazuo Gao, Chenggang Zhao, Xu~Sun, and Damai Dai.
\newblock Auxiliary-loss-free load balancing strategy for mixture-of-experts.
\newblock \emph{arXiv preprint arXiv:2408.15664}, 2024{\natexlab{a}}.

\bibitem[Wang et~al.(2024{\natexlab{b}})Wang, Ma, Zhang, Ni, Chandra, Guo, Ren, Arulraj, He, Jiang, Li, Ku, Wang, Zhuang, Fan, Yue, and Chen]{mmlu-pro}
Yubo Wang, Xueguang Ma, Ge~Zhang, Yuansheng Ni, Abhranil Chandra, Shiguang Guo, Weiming Ren, Aaran Arulraj, Xuan He, Ziyan Jiang, Tianle Li, Max Ku, Kai Wang, Alex Zhuang, Rongqi Fan, Xiang Yue, and Wenhu Chen.
\newblock Mmlu-pro: {A} more robust and challenging multi-task language understanding benchmark.
\newblock In Amir Globersons, Lester Mackey, Danielle Belgrave, Angela Fan, Ulrich Paquet, Jakub~M. Tomczak, and Cheng Zhang (eds.), \emph{Advances in Neural Information Processing Systems 38: Annual Conference on Neural Information Processing Systems 2024, NeurIPS 2024, Vancouver, BC, Canada, December 10 - 15, 2024}, 2024{\natexlab{b}}.
\newblock URL \url{http://papers.nips.cc/paper\_files/paper/2024/hash/ad236edc564f3e3156e1b2feafb99a24-Abstract-Datasets\_and\_Benchmarks\_Track.html}.

\bibitem[Wei et~al.(2023)Wei, Luan, Liu, Dong, and Wang]{cmath}
Tianwen Wei, Jian Luan, Wei Liu, Shuang Dong, and Bin Wang.
\newblock {CMATH:} can your language model pass chinese elementary school math test?
\newblock \emph{CoRR}, abs/2306.16636, 2023.
\newblock \doi{10.48550/ARXIV.2306.16636}.
\newblock URL \url{https://doi.org/10.48550/arXiv.2306.16636}.

\bibitem[Xiao et~al.(2023)Xiao, Tian, Chen, Han, and Lewis]{xiao2023efficient}
Guangxuan Xiao, Yuandong Tian, Beidi Chen, Song Han, and Mike Lewis.
\newblock Efficient streaming language models with attention sinks.
\newblock \emph{arXiv preprint arXiv:2309.17453}, 2023.

\bibitem[Xiao et~al.(2024)Xiao, Tang, Zuo, Guo, Yang, Tang, Fu, and Han]{xiao2024duoattention}
Guangxuan Xiao, Jiaming Tang, Jingwei Zuo, Junxian Guo, Shang Yang, Haotian Tang, Yao Fu, and Song Han.
\newblock Duoattention: Efficient long-context llm inference with retrieval and streaming heads.
\newblock \emph{arXiv preprint arXiv:2410.10819}, 2024.

\bibitem[Yang et~al.(2025)Yang, Li, Yang, Zhang, Hui, Zheng, Yu, Gao, Huang, Lv, et~al.]{yang2025qwen3}
An~Yang, Anfeng Li, Baosong Yang, Beichen Zhang, Binyuan Hui, Bo~Zheng, Bowen Yu, Chang Gao, Chengen Huang, Chenxu Lv, et~al.
\newblock Qwen3 technical report.
\newblock \emph{arXiv preprint arXiv:2505.09388}, 2025.

\bibitem[Zadouri et~al.(2025)Zadouri, Strauss, and Dao]{zadouri2025hardware}
Ted Zadouri, Hubert Strauss, and Tri Dao.
\newblock Hardware-efficient attention for fast decoding.
\newblock \emph{arXiv preprint arXiv:2505.21487}, 2025.

\bibitem[Zellers et~al.(2019)Zellers, Holtzman, Bisk, Farhadi, and Choi]{hellaswag}
Rowan Zellers, Ari Holtzman, Yonatan Bisk, Ali Farhadi, and Yejin Choi.
\newblock Hellaswag: Can a machine really finish your sentence?
\newblock In Anna Korhonen, David~R. Traum, and Llu{\'{\i}}s M{\`{a}}rquez (eds.), \emph{Proceedings of the 57th Conference of the Association for Computational Linguistics, {ACL} 2019, Florence, Italy, July 28- August 2, 2019, Volume 1: Long Papers}, pp.\  4791--4800. Association for Computational Linguistics, 2019.
\newblock \doi{10.18653/V1/P19-1472}.
\newblock URL \url{https://doi.org/10.18653/v1/p19-1472}.

\bibitem[Zhong et~al.(2024)Zhong, Cui, Guo, Liang, Lu, Wang, Saied, Chen, and Duan]{agieval}
Wanjun Zhong, Ruixiang Cui, Yiduo Guo, Yaobo Liang, Shuai Lu, Yanlin Wang, Amin Saied, Weizhu Chen, and Nan Duan.
\newblock Agieval: {A} human-centric benchmark for evaluating foundation models.
\newblock In Kevin Duh, Helena G{\'{o}}mez{-}Adorno, and Steven Bethard (eds.), \emph{Findings of the Association for Computational Linguistics: {NAACL} 2024, Mexico City, Mexico, June 16-21, 2024}, pp.\  2299--2314. Association for Computational Linguistics, 2024.
\newblock \doi{10.18653/V1/2024.FINDINGS-NAACL.149}.
\newblock URL \url{https://doi.org/10.18653/v1/2024.findings-naacl.149}.

\end{thebibliography}
\bibliographystyle{KHA}

% Appendix
\clearpage
\appendix
\section{Appendix}
\subsection{Comparison of Interactive-heads Attention Mechanisms}
\label{sec:comparison}
We provide a comprehensive comparison of our knocking-heads attention with existing interactive-head mechanisms across multiple dimensions, as summarized in Table~\ref{tab:comparison}. Our analysis encompasses three representative approaches: talking-heads attention~\citep{shazeer2020talking}, collaborated multi-head attention (CollabHead)~\citep{cordonnier2020multi}, and mixture-of-head (MoH)~\citep{jin2024moh}.

The compared methods employ fundamentally different interaction strategies. Talking-heads attention uses learnable transition matrices to directly combine attention weights across heads, achieving strong interaction but with high complexity and FlashAttention incompatibility. MoH learns routers to select heads for different tokens, promoting head specialization but lacking direct head interaction. CollabHead enables head interaction by replacing individual head transformation matrices with one large shared matrix across all heads, which compromises head specification and increases training FLOPs due to the enlarged shared matrix. Our knocking-heads mechanism achieves head interaction through lightweight feature-sharing modules inserted into existing attention variants, maintaining both strong interaction and head specification with minimal overhead.

\begin{table*}[!h]
\centering
\small
\caption{Comparison of interactive-heads attention mechanisms: talking-heads~\citep{shazeer2020talking}, collaborated multi-head~\citep{cordonnier2020multi}, mixture-of-head~\citep{jin2024moh}, and our knocking-heads attention. ``compute control" is about FLOPs, ``compatibility" includes attention variants and FlashAttention support, and ``training stability" indicates loss spike frequency.}
\resizebox{0.9\linewidth}{!}{\begin{tabular}{@{}lccccccc@{}}
\toprule
& \makecell{Interaction\\Method} & \makecell{Head-\\Interaction} & \makecell{Head-\\Specifiction} & \makecell{Compa-\\tibility} & \makecell{Compute\\Control} & \makecell{Param\\Control} & \makecell{Training\\Stability}\\ 
\midrule
Talking-heads    & Mixing & \textcolor{darkgreen}{\ding{51}}Strong& \textcolor{darkgreen}{\ding{51}}Strong& \textcolor{darkred}{\ding{55}} & \textcolor{darkred}{\ding{55}} &\textcolor{darkgreen}{\ding{51}} & unknown\\
Collaborated-heads     & Sharing & \textcolor{darkgreen}{\ding{51}}Strong& \textcolor{darkred}{\ding{55}}Weak& \textcolor{darkred}{\ding{55}} & \textcolor{darkred}{\ding{55}} &\textcolor{darkgreen}{\ding{51}} & unknown\\
Mixture-of-head  & Re-weight & \textcolor{darkred}{\ding{55}}Weak& \textcolor{darkgreen}{\ding{51}}Strong& \textcolor{darkgreen}{\ding{51}} &\textcolor{darkgreen}{\ding{51}} &\textcolor{darkgreen}{\ding{51}} & unknown\\
\rowcolor{gray!10} Knocking-heads(Ours)   & Sharing & \textcolor{darkgreen}{\ding{51}}Strong& \textcolor{darkgreen}{\ding{51}}strong& \textcolor{darkgreen}{\ding{51}} &\textcolor{darkgreen}{\ding{51}} &\textcolor{darkgreen}{\ding{51}} & \textcolor{darkgreen}{\ding{51}}\\
\bottomrule
\end{tabular}}
\label{tab:comparison}
\end{table*}

\subsection{Evaluation Benchmark}
\label{sec:benchmark}
We assess model performance using a comprehensive benchmark spanning multiple downstream tasks, which collectively measure different aspects of model competence. The evaluation framework is organized into distinct task categories, including: (a) General Knowledge (\eg ARC~\citep{arc}, AGIEval~\citep{agieval}, PIQA~\citep{piqa}, HellaSwag~\citep{hellaswag}) (b) Language Understanding (\eg RACE~\citep{race}) (c) Professional Knowledge (\eg MMLU~\citep{mmlu}, CMMLU~\citep{cmmlu}, MMLU-Pro~\citep{mmlu-pro}, GPQA~\citep{gpqa}, C-Eval~\citep{ceval}, CommonsenseQA~\citep{talmor2018commonsenseqa})) (d) Math (\eg GSM8K~\citep{gsm8k}, MATH~\citep{math}, CMATH~\citep{cmath} (e) Code (\eg HumanEval-plus~\citep{evalplus}, MBPP~\citep{mbpp}, MBPP-Plus~\citep{evalplus}.

\end{document}